\documentclass[10pt,journal,compsoc]{IEEEtran}

\usepackage{amsmath}
\usepackage{diagbox}
\usepackage{graphics}
\usepackage{epsfig}
\usepackage{threeparttable}
\usepackage{xspace}
\usepackage{siunitx}
\usepackage{graphicx}
\usepackage{amssymb}
\usepackage{threeparttable}
\usepackage{color}
\usepackage[normalem]{ulem}
\usepackage{multirow}
\usepackage{float}
\usepackage{amsfonts}
\usepackage{bm}
\usepackage{array}
\usepackage[table]{xcolor}
\usepackage{colortbl}
\usepackage{diagbox}
\usepackage{rotating}
\usepackage{booktabs}
\usepackage{overpic}
\usepackage{enumitem}
\usepackage{colortbl}
\usepackage{algorithm}
\usepackage{algorithmic}
\usepackage{cite}
\usepackage{dblfloatfix}
\usepackage{hyphenat}

\usepackage{pifont}
\usepackage[perpage,symbol*]{footmisc}
\DefineFNsymbols{circled}{{\ding{192}}{\ding{193}}{\ding{194}}
	{\ding{195}}{\ding{196}}{\ding{197}}{\ding{198}}{\ding{199}}{\ding{200}}{\ding{201}}}
\setfnsymbol{circled}

\newcommand\myfootnotestyle[1]{\ifcase#1 \or \ding{182}\or \ding{183}\or
	\ding{184}\or \ding{185}\or \ding{186}\or \ding{187}%
	\or \ding{188}\or \ding{189}\or \ding{190}\or \ding{191}\else *\fi\relax}

\usepackage[pagebackref=false,breaklinks=true,colorlinks,bookmarks=false]{hyperref}

\definecolor{mygray}{gray}{.85}
\definecolor{mygray1}{gray}{.7}
\definecolor{mygray2}{gray}{.93}

\newcommand{\pa}[1]{\noindent \emph{\textbf{#1}}}

\makeatletter
\newcommand{\thickhline}{%
	\noalign {\ifnum 0=`}\fi \hrule height 1pt
	\futurelet \reserved@a \@xhline
}
\makeatother

\makeatletter
\DeclareRobustCommand\onedot{\futurelet\@let@token\@onedot}
\def\@onedot{\ifx\@let@token.\else.\null\fi\xspace}

\def\eg{\emph{e.g}\onedot} 
\def\ie{\emph{i.e}\onedot}

\makeatother

\DeclareSymbolFont{extraup}{U}{zavm}{m}{n}
\DeclareMathSymbol{\varheart}{\mathalpha}{extraup}{86}
\DeclareMathSymbol{\vardiamond}{\mathalpha}{extraup}{87}

\begin{document}
	\title{Image Segmentation in Foundation Model Era: A  Survey}
	
	\author{Tianfei Zhou, Wang Xia, Fei Zhang, Boyu Chang, Wenguan Wang, Ye Yuan, \\Ender Konukoglu, Daniel Cremers 
		\IEEEcompsocitemizethanks{
			\IEEEcompsocthanksitem T. Zhou, W. Xia, B. Chang and Y. Yuan are with Department of Computer Science, Beijing Institute of Technology, China.
			\IEEEcompsocthanksitem F. Zhang is with Department of Electronic Engineering, Shanghai Jiao Tong University, China.
			\IEEEcompsocthanksitem W. Wang is  with CCAI, Zhejiang University, China. 
			\IEEEcompsocthanksitem E. Konukoglu is with Computer Vision Lab, ETH Zurich, Switzerland.
			\IEEEcompsocthanksitem D. Cremers is with the Department of
			Computer Science, Technical University of Munich, Germany.
			\\ \\
			E-mail: tfzhou@bit.edu.cn (Tianfei Zhou)
		}
	}

	
	\IEEEtitleabstractindextext{
		\begin{abstract}
			Image segmentation is a long-standing challenge in computer vision, studied continuously over several decades, as evidenced by seminal algorithms such as N-Cut, FCN, and MaskFormer. With the advent of foundation models (FMs), contemporary segmentation methodologies have embarked on a new epoch by either adapting FMs (\eg, CLIP, Stable Diffusion, DINO) for image segmentation or developing dedicated  segmentation foundation models (\eg, SAM, SAM2). These approaches not only deliver superior segmentation performance, but also herald newfound segmentation capabilities previously unseen in deep learning context. However, current research in image segmentation lacks a detailed analysis of  distinct characteristics, challenges, and solutions associated with these advancements. This survey seeks to fill this gap by providing a thorough review of  cutting-edge research centered around FM-driven image segmentation. We investigate two basic  lines of research  -- generic image segmentation (\ie, semantic segmentation, instance segmentation, panoptic segmentation), and promptable image segmentation (\ie, interactive segmentation, referring segmentation, few-shot segmentation) -- by delineating their respective task settings, background concepts, and key challenges. Furthermore, we provide insights into the emergence of segmentation knowledge from FMs like CLIP, Stable Diffusion, and DINO. An exhaustive overview of over 300 segmentation approaches is provided to encapsulate the breadth of current research efforts. Subsequently, we  engage in a discussion of open issues and potential avenues for future research. We envisage that this fresh, comprehensive, and systematic survey catalyzes the evolution of  advanced image segmentation systems. A public website is created to continuously track developments in this fast advancing field: \url{https://github.com/stanley-313/ImageSegFM-Survey}.

		\end{abstract}
		}
	
	\maketitle
	\IEEEdisplaynontitleabstractindextext
	\IEEEpeerreviewmaketitle

	\IEEEraisesectionheading{\section{Introduction}\label{sec:introduction}}
	\label{sec:intro}

	\IEEEPARstart{I}{mage} segmentation has been, and still is, an important and challenging research field in computer vision, with its  aim to partition pixels into distinct groups.  It constitutes an initial step in achieving higher-order goals including physical scene understanding, reasoning over visual commonsense, perceiving social affordances, and has widespread applications in   domains like   autonomous driving, medical image analysis,  automated surveillance, and image editing.
	
	The task has garnered extensive attention over decades, resulting in a plethora of algorithms in the literature, \textit{ranging from  traditional, non-deep learning methods} such as thresholding \cite{otsu1975threshold, yanowitz1989new}, histogram mode seeking \cite{ohta1980color,deng1999color}, region growing and merging \cite{adams1994seeded,nock2004statistical}, spatial clustering \cite{shi2000normalized}, energy diffusion \cite{ma1997edge}, superpixels \cite{achanta2012slic},  conditional and Markov random fields \cite{he2004multiscale}, 
	\textit{to more advanced, deep learning methods}, \eg, FCN-based \cite{long2015fully,wang2020deep,li2022deep,liu2023tripartite, ronneberger2015u,zhou2024cross,chen2014semantic,chen2017deeplab,chen2017rethinking,chen2018encoder} and particularly the DeepLab family \cite{chen2014semantic,chen2017deeplab,chen2017rethinking,chen2018encoder}, RNN-based \cite{byeon2015scene},  Transformer-based \cite{cheng2021per,cheng2022masked,liang2023clustseg,jain2023oneformer,zhou2022rethinking,zheng2021rethinking,liang2023local}, and the R-CNN family \cite{girshick2015fast,ren2016faster,he2017mask}. These approaches have shown remarkable performance and robustness  across all critical segmentation fields, \eg, semantic, instance, and panoptic segmentation.  Yet, the exploration of image segmentation continues beyond these advancements. 
	
	Foundation Models (FMs)  \cite{bommasani2021opportunities} have emerged as transformative technologies in recent years, reshaping our understanding of core domains in artificial intelligence (AI) including natural language processing \cite{zhao2023survey}, computer vision \cite{awais2023foundational}, and many other interdisciplinary areas \cite{latif2023sparks,thirunavukarasu2023large,jin2023large}. Notable examples include large language models (LLMs) like GPT-3 \cite{brown2020language} and GPT-4 \cite{achiam2023gpt}, multimodal large language models (MLLMs) like Flamingo \cite{alayrac2022flamingo} and Gemini \cite{team2023gemini}, and diffusion  models (DMs) like Sora \cite{openai2024sora}  and Stable Diffusion (SD) \cite{rombach2022high}. These models, distinguished by their immense scale and complexity, have exhibited emergent capabilities \cite{schaeffer2024emergent,wei2022emergent} to tackle a wide array of intricate tasks with notable efficacy and efficiency. Meanwhile, they have  unlocked new  possibilities, such as generating chains of reasoning \cite{wei2022chain}, offering human-like responses in dialogue scenarios \cite{brown2020language}, creating realistic-looking videos \cite{openai2024sora}, and synthesizing novel programs \cite{chen2021evaluating}. The advent of GPT-4 and Sora has sparked considerable excitement within the AI community to fulfill artificial general intelligence (AGI) \cite{OpenAI-blog-2023-Planning}.
	
	In the era dominated by FMs, image segmentation has undergone significant evolution, marked by distinct features uncommon in the preceding research era. To underscore the motivation behind our survey, we highlight several characteristics exemplifying this transformation:

	\noindent$\spadesuit$~FM technology has led to the emergence of segmentation generalists. Unlike traditional frameworks (\eg, FCN, Mask R-CNN),  contemporary segmentation models have become  promptable, \ie, generate a mask (akin to an answer in LLMs) based on a handcrafted prompt  specifying what to segment in an image. The LLM-like promptable interface leads to a significant enhancement of task generality of segmentors, enabling them to rapidly adapt to various existing and new segmentation tasks, in a zero-shot (\eg, SAM \cite{kirillov2023segment}, SEEM \cite{zou2024segment}) or few-shot (\eg, SegGPT \cite{wang2023seggpt}) manner. Note that these promptable models  markedly differ from earlier universal models \cite{cheng2022masked,liang2023clustseg,cheng2021per,jain2023oneformer}, which remain limited to a fixed set of predetermined tasks, \eg, joint  semantic, instance, and panoptic segmentation, with a closed vocabulary.
	
	\noindent$\varheart$~Training-free segmentation has recently emerged as a burgeoning research area \cite{zhou2022extract,wang2023sclip,barsellotti2024training,wang2023diffusion,caron2021emerging,oquab2024dinov2}. It aims to extract segmentation  knowledge  from pre-trained FMs, marking a departure from established learning paradigms, such as
	supervised, semi-supervised, weakly supervised,  and self-supervised learning. Recent studies highlight that segmentation masks can be derived effortlessly from attention maps or internal representations within models like CLIP, Stable Diffusion or DINO/DINOv2, even though they were not originally designed for segmentation purposes. 

	\noindent$\clubsuit$~There is a notable trend towards integrating LLMs into segmentation systems to harness their reasoning capabilities and world knowledge \cite{rasheed2024glamm,lai2024lisa,ren2024pixellm,zhang2024psalm}. The LLM-powered segmentors possess the capacity to read, listen, and even reason to ground real-world, abstract linguistic queries into specific pixel regions. While previous efforts have explored similar capabilities in tasks  such as referring segmentation \cite{kazemzadeh2014referitgame}, these methods are limited in handling basic queries like ``\textit{the front-runner}''. In  contrast, LLM-powered segmentors can  adeptly manage  more complicated queries   like ``\textit{who will win the race?}''. This capability represents a notable advancement towards developing more intelligent vision systems.
	
	\noindent$\vardiamond$~Generative models, particularly text-to-image diffusion models, garner increasing  attention in recent image segmentation research.   It has been observed that DMs implicitly learn meaningful object groupings and semantics during the text-to-image generation process \cite{baranchuk2022label}, functioning as strong unsupervised representation learners.  This motivates  a stream of works to directly decode the latent code of pre-trained DMs into segmentation masks,   in either a label-efficient  or completely unsupervised manner \cite{baranchuk2022label,tian2023diffuse}. Moreover, some efforts extend the inherent denoising diffusion process in DMs to segmentation, by approaching image segmentation  from  an image-conditioned mask generation perspective \cite{ji2023ddp,chen2023generative,wang2023dformer}.
	
	In light of these features, we found that  most existing surveys in the field \cite{cheng2001color,zhang1996survey,minaee2021image} are  now outdated -- one of the latest surveys \cite{minaee2021image} was published in 2021 and focuses only on  semantic and instance segmentation. This  leaves a notable gap in capturing recent  FM-based  approaches.
	
	\vspace{2mm}
	\pa{Our Contributions.} To fill the gap, we offer an exhaustive and timely overview to examine \textit{how foundation models are transforming the field of image segmentation}.
	This survey marks the {first} comprehensive exploration of  recent  image segmentation approaches that are built upon famous FMs, such as CLIP \cite{radford2021learning}, Stable Diffusion \cite{rombach2022high}, DINO \cite{caron2021emerging}/DINOv2 \cite{oquab2024dinov2}, SAM \cite{kirillov2023segment} and LLMs/MLLMs \cite{yin2023survey}. It spans the breadth of the field and delves into the nuances of individual methods, thereby providing the reader with a thorough and up-to-date understanding of this topic. Beyond this, we elucidate open questions and potential directions to illuminate the path forward in this key field. 

	   \begin{figure}[t]
		\centering
		\includegraphics[width=0.99\linewidth]{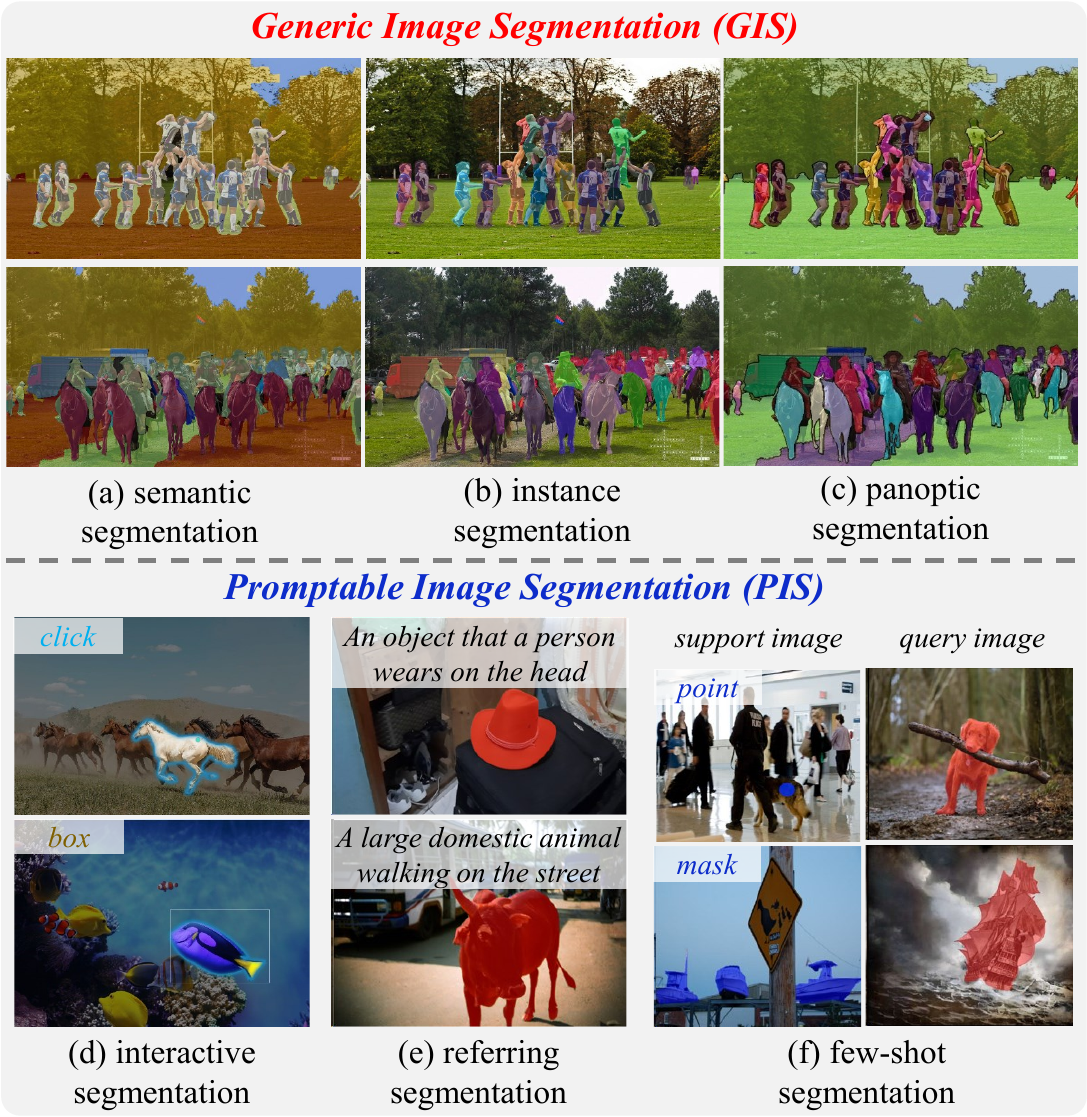}
		\caption{\textbf{Image segmentation tasks reviewed in this survey}. Generic image segmentation: (a) semantic segmentation, (b) instance segmentation, (c) panoptic segmentation; Promptable image segmentation: (d) interactive segmentation, (e) referring segmentation,  (f) few-shot segmentation.}
		\label{fig:1}
	\end{figure}

	\begin{figure*}[t]
	\centering
	\includegraphics[width=0.99\linewidth]{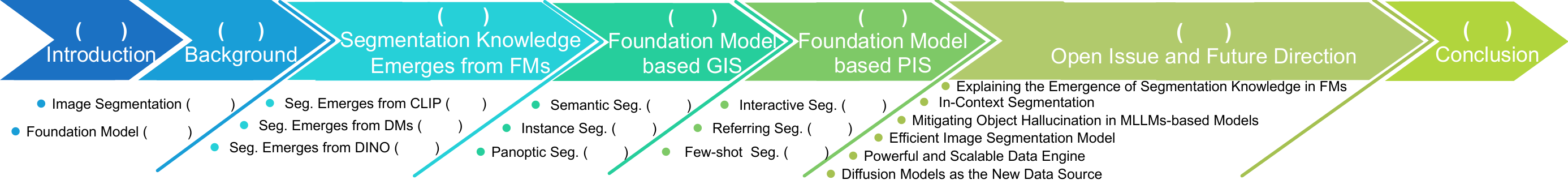}
        \put(-482, 48){\scriptsize \S\ref{sec:introduction}}
        \put(-436.8, 48){\scriptsize \S\ref{sec:2}}
        \put(-448, 24){\tiny \S\ref{sec:2.1}}
        \put(-462, 15.6){\tiny \S\ref{sec:2.2}}
        \put(-364, 52.8){\scriptsize \S\ref{sec:3}}
        \put(-364.7, 25){\tiny \S\ref{sec:3.1}}
        \put(-373, 17.5){\tiny \S\ref{sec:3.2}}
        \put(-380.8, 10.5){\tiny \S\ref{sec:3.3}}
        \put(-290, 51.6){\scriptsize \S\ref{sec:4}}
        \put(-298, 24){\tiny \S\ref{sec:4.1}}
        \put(-309.6, 16){\tiny \S\ref{sec:4.2}}
        \put(-319, 8.5){\tiny \S\ref{sec:4.3}}
        \put(-227, 51.8){\scriptsize \S\ref{sec:5}}
        \put(-234, 24){\tiny \S\ref{sec:5.1}}
        \put(-246, 16){\tiny \S\ref{sec:5.2}}
        \put(-254, 9){\tiny \S\ref{sec:5.3}}
        \put(-123, 48){\scriptsize \S\ref{sec:6}}
        \put(-29, 48){\scriptsize \S\ref{sec:7}}
	\caption{Overview of this survey.}
	\label{fig:overview}
	\end{figure*}
	
	\vspace{2mm}
	\pa{Related Surveys and Differences.} In the past decade, many surveys have studied image segmentation from various perspectives. For example, \cite{zaitoun2015survey} reviews  region- and boundary-based segmentation methods in 2015. With the transition to the deep learning era, a series of works \cite{ghosh2019understanding,minaee2021image,liu2019recent,lateef2019survey,shen2023survey,csurka2022semantic} has summarized progress in generic segmentation tasks like semantic, instance and panoptic segmentation. A recent study \cite{zhu2024survey} focuses on the specific task of open-vocabulary segmentation, while \cite{li2024transformer} only studies Transformer-based segmentation. Other studies delve into crucial aspects of image segmentation, such as evaluation protocols \cite{wang2020image} or loss functions \cite{jadon2020survey}. In addition, there exist surveys that focus on  segmentation techniques in specialized domains, \eg, video \cite{zhou2022survey},  medical imaging \cite{wang2022medical,noble2006ultrasound}. 

	Given the accelerated evolution of FMs, there has been a surge of surveys that elucidate the fundamental principles and pioneering efforts in LLMs \cite{zhao2023survey}, MLLMs \cite{yin2023survey}, DMs \cite{croitoru2023diffusion}. However, conspicuously absent from these works is a discussion on the role of FMs in advancing image segmentation.  
	The survey most relevant to ours is \cite{zhang2023comprehensive}, which offers an extensive review of  recent developments related to SAM \cite{kirillov2023segment}.  SAM represents  a groundbreaking contribution to the segmentation field,  making \cite{zhang2023comprehensive} a valuable resource. However, within the broader context of FMs, SAM is just one among many; thus, the scope of \cite{zhang2023comprehensive} is still limited in encompassing the entirety of progress in segmentation field.
	
	Unlike prior surveys, our work stands apart in its exclusive focus on the contributions of  FMs to image segmentation, and fills an existing gap in the current research landscape.  We document the latest techniques, and spotlight major trends, and envision prospective research trajectories which will aid researchers in staying abreast of advances in image segmentation and  accelerate progress in the field.
	

	\vspace{2mm}
	\pa{Survey Organization.} Fig.~\ref{fig:overview} shows the structure of this survey. Section \S\ref{sec:2} presents essential background on image segmentation and FMs. \S\ref{sec:3} highlights the emergency of segmentation knowledge from existing FMs.  \S\ref{sec:4}  and \S\ref{sec:5} review the most important FM-based image segmentation methods, mainly from the past three years.  \S\ref{sec:6} raises open issues and future directions. We conclude the paper in \S\ref{sec:7}.

	\section{Background}\label{sec:2}
	In this section, we first present a unified formulation of image segmentation tasks and categorize research directions  in \S\ref{sec:2.1}. Then, we provide a concise background overview of prominent FMs in  \S\ref{sec:2.2}.
	
	\subsection{Image Segmentation}\label{sec:2.1}
	
	\subsubsection{A Unified Formulation}
	
	The central goal of the paper is to investigate the contributions  of FMs to modern image segmentation technology. To this end, we first introduce a unified mathematical formulation applicable to various segmentation tasks. Concretely, denote $\bm{\mathcal{X}}$ and $\bm{\mathcal{Y}}$ as the input space and output segmentation space, respectively. An image segmentation  solution  seeks to learn an ideal mapping function $f$:
	\begin{equation}\small\label{eq:1}
		\begin{aligned}
			f:~\bm{\mathcal{X}}\mapsto\bm{\mathcal{Y}}, 
			~~\text{where}~~\bm{\mathcal{X}}=\mathcal{I}\times\mathcal{P}, ~~\bm{\mathcal{Y}}=\mathcal{M}\times\mathcal{C}.
		\end{aligned}
	\end{equation}
	Here $f$ is typically instantiated as a neural network. The input space $\bm{\mathcal{X}}$ is decomposed as $\mathcal{I}\times\mathcal{P}$, where $\mathcal{I}$ represents an image domain (comprising  solely  a single image $I$), and  $\mathcal{P}$ refers to a collection of  prompts, which is exclusively employed in certain segmentation tasks. The output space is $\bm{\mathcal{Y}}\!=\!\mathcal{M}\times\mathcal{C}$, which encompasses a set of segmentation mask $\mathcal{M}$ and a vocabulary $\mathcal{C}$ of semantic categories  associated with these masks. Eq.~\ref{eq:1} furnishes a structured framework for understanding image segmentation, wherein a neural network is trained to map an input image, along with potentially user-specified prompts, to segmentation masks as well as corresponding semantic categories. Based on Eq.~\ref{eq:1}, we subsequently build a taxonomy for image segmentation.

	\subsubsection{Image Segmentation Category}
	
	According to whether $P$ is provided, we categorize image segmentation methods into two  classes (Fig.~\ref{fig:1}): \textit{generic image segmentation (GIS)} and \textit{promptable image segmentation (PIS)}.

	\noindent~$\bullet$~\textbf{\textit{Generic Image Segmentation}.} GIS aims to  segment an image  into distinct regions, each associated with a semantic category or an object. In GIS, the input space comprises solely the image, \ie, $\bm{\mathcal{X}}\equiv\mathcal{I}$, indicating $\mathcal{P}\!=\!\emptyset$. Based on the definition of output space $\bm{\mathcal{Y}}$, GIS methods can be further categorized into three major types: \textit{(i) semantic segmentation} (Fig.~\ref{fig:1}a) aims to identify and label each pixel with  a semantic class in $\mathcal{C}$. \textit{(ii) instance segmentation} (Fig.~\ref{fig:1}b) involves grouping pixels that belong to the same semantic class into separate object instances.  \textit{(iii) panoptic segmentation} (Fig.~\ref{fig:1}c) integrates  semantic and instance segmentation to predict per-pixel class and instance labels, and  is able to  provide a comprehensive scene parsing. 
	
	Furthermore, based on whether the testing vocabulary  $\mathcal{C}_{test}$ includes novel classes absent from the training vocabulary $\mathcal{C}_{train}$,  the three tasks are studied under two  settings: \textit{closed-vocabulary} (\ie, $\mathcal{C}_{train}\!\equiv\!\mathcal{C}_{test}$) and \textit{open-vocabulary} (\ie, $\mathcal{C}_{train}\!\subset\!\mathcal{C}_{test}$) segmentation. Notably, the closed-vocabulary setup has been extensively studied over the past decade. However, its open-vocabulary counterpart is still in its infancy and has garnered attention only in recent years, particularly with the advent of FMs.

	\noindent~$\bullet$~\textbf{\textit{Promptable Image Segmentation}.} PIS extends GIS by additionally incorporating a set of prompts $\mathcal{P}$, specifying the target to  segment. In general, PIS methods only generate segmentation masks closely related to the prompts and do not directly predict classes. While the term ``prompt'' is relatively new, PIS has been studied for many years. Depending upon the prompt type, PIS methods can be  grouped into the following categories: \textit{(i) interactive segmentation} (Fig.~\ref{fig:1}d) aims to segment out specific objects or parts according to user input, often provided through clicks, scribbles, boxes, or polygons, thus $\mathcal{P}\!=\!\{click, scribble, box, polygon\}$ are visual prompts; \textit{(ii) referring segmentation} (Fig.~\ref{fig:1}e) entails extracting the corresponding region referred by a linguistic phrase, thus $\mathcal{P}\!=\!\{linguistic~phrase\}$ refers to textual prompts; \textit{(iii) few-shot segmentation} (Fig.~\ref{fig:1}f) targets at segmenting novel objects in given query image with a few annotated support images, \ie, $\mathcal{P}\!=\!\{(image, mask)\}$ refers to a collection  of image-mask pairs. While great progress has been made in these segmentation challenges, previous studies address various prompt types independently. In sharp contrast,  FM-based methods aim to integrate them into a unified framework. Moreover, \textit{in-context segmentation} has emerged as a novel few-shot segmentation task. 
	
	\subsubsection{Learning Paradigms for Image Segmentation}
	
	Several prevalent learning strategies are employed to approximate the function $f$ in Eq.~\ref{eq:1}. 
	\textit{(i) Supervised learning}: modern image segmentation methods are generally learned in a fully supervised manner, necessitating a collection of training images and their desired outputs, \ie per-pixel annotations. 
	\textit{(ii) Unsupervised learning}: in the absence of explicit annotated supervision, the task of approximating $f$ falls under unsupervised learning. Most existing unsupervised learning-based image segmentation models utilize self-supervised techniques,  training networks with automatically-generated pseudo labels derived from image data.  
	\textit{(iii) Weakly-supervised learning}: in this case, supervision information may be  inexact, incomplete or inaccurate. For inexact supervision, labels are typically acquired from a more easily annotated domain (\eg, image tag, bounding box, scribble). In the case of incomplete supervision, labels are  provided   for only a subset of training images.  Inaccurate supervision entails per-pixel annotations for all training images, albeit with the presence of noise. 
	\textit{(iv) Training free}: in addition to the aforementioned strategies, a novel paradigm -- \textit{training-free segmentation} -- has gained  attention in the FM era, aiming to extract segmentation directly from pre-trained FMs, without involving any model training.

%
%
	
	\subsection{Foundation Model}\label{sec:2.2}
	
	FMs are initially elucidated  in \cite{bommasani2021opportunities} as \textit{``any model that is trained on broad data (generally using self-supervision at scale) that can be adapted to a wide range of downstream tasks''.}  The term `foundation' is used to underscore  critically central and incomplete character of FMs: homogenization of the methodologies across research communities and emergence of new capabilities. While the basic ingredients of the FMs, such as deep neural networks and self-supervised learning, have been around for many years, the paradigm shift towards FMs is significant because the emergence and homogenization allow replacing narrow task-specific models with more generic task-agnostic models that are not strongly tied to a particular task or domain. In the subsequent subsections, we provide a brief review of language  (\S\ref{sec:2.3.1}) and vision foundation models (\S\ref{sec:2.3.2}). Notably, we only focus on topics relevant to this survey, and direct interested readers to \cite{liu2023towards,zhao2023survey} for more comprehensive discussions.
	
	\subsubsection{Language Foundation Model} \label{sec:2.3.1}
	
	\noindent~$\bullet$~\textbf{\textit{Large Language Models (LLMs).}} Language modeling  is one of the primary approaches to advancing language intelligence of machines. In general, it aims to model the generative likelihood of word sequences, so as to \textit{predict the probabilities of future tokens}. In the past two decades, language modeling has evolved from the earliest  statistical language models (SLMs) to neural language models (NLMs), then to small-sized pre-trained language models (PLMs), and finally to nowadays LLMs \cite{zhao2023survey}. As enlarged PLMs (in terms of model size, data size and  training compute), LLMs not only achieve a significant zero-shot performance improvement (even in some cases matching  finetuned models), but also show strong reasoning capabilities across various domains, \eg, code writing \cite{chen2021evaluating}, math problem solving \cite{lewkowycz2022solving}.  A remarkable application of LLMs is \textit{ChatGPT}, which has attracted widespread attention  and  transformed the way we interact with AI technology.
	
	\noindent~$\bullet$~\textbf{\textit{Multimodal Large Language Models (MLLMs).}} MLLMs \cite{yin2023survey} are multimodal  extensions of LLMs by bringing together the \textit{reasoning} power of LLMs with the capability to process non-textual modalities (\eg, vision, audio).   MLLMs represent the next level of LLMs. On one hand, multimodal perception  is a natural way for knowledge acquisition and interaction with the real world, and thus serves as a fundamental component for achieving AGI; on the other hand, the multimodal extension expands the potential of pure language modeling to more complex tasks in, \eg, robotics and autonomous driving.
	

	\subsubsection{Visual Foundation Model}  \label{sec:2.3.2}
	
	\noindent~$\bullet$~\textbf{\textit{Contrastive Language-Image Pre-training (CLIP).}} CLIP \cite{radford2021learning} embodies a language-supervised vision  model trained on 400M image-text pairs sourced from the Internet. The model has an encoder-only architecture, consisting of separate encoders  for image and text encoding.  It is  trained by an image-text contrastive learning objective:
		\vspace{-3pt}
	\begin{equation}\small\label{eq:clip}
		\mathcal{L}_{i2t} = -\log \left[ \frac{\exp(\text{sim} (\bm{x}_k, \bm{t}_k)/\tau)} {\sum_{j=1}^J \exp(\text{sim}(\bm{x}_k, \bm{t}_j)/\tau)} \right].
			\vspace{-3pt}
	\end{equation}
	where $(\bm{x}_k, \bm{t}_k)$ denotes the image and text embeddings of the $k$-th image-text example $(x_k, t_k)$. $J$ and $\tau$ indicate the  number of examples and softmax temperature, respectively. The loss $\mathcal{L}_{i2t\!}$ maximizes agreement between the embeddings of matching image and text pairs while minimizing it for non-matching pairs. In practice, text-image contrastive loss is calculated similarly, and the model is  trained by a joint loss: $\mathcal{L}_\text{contrast}\!=\! \mathcal{L}_{i2t} \!+\! \mathcal{L}_{t2i}$.  ALIGN \cite{jia2021scaling} is a follow-up work that harnesses $\mathcal{L}_\text{contrast}$ for visual representation learning. It simplifies the costly data curation process in CLIP, and succeeds to further scale up representation learning with a noisy dataset of over one billion image-text pairs. Both CLIP and ALIGN  acquire semantically-rich visual concepts and demonstrate impressive transferability in recognizing novel categories, leading to increased adoption for tackling  zero-shot and open-vocabulary recognition tasks.

	\noindent~$\bullet$~\textbf{\textit{Diffusion Models (DMs)}}. DMs  are a family of generative models that are Markov chains trained with variational inference. They have demonstrated remarkable potential in creating visually realistic samples, and set the current state-of-the-art in generation tasks. The milestone work, denoising diffusion probabilistic model (DDPMs) \cite{ho2020denoising}, was published in 2020 and have sparked an exponentially increasing interest in the generative AI community afterwards. DDPMs are defined as a parameterized Markov chain, which generate data from Gaussian noise within finite transitions during inference. Its training encompasses two interconnected processes. \textbf{(i) Forward pass} maps a data distribution $\bm{z}_0\!\sim\!q(\bm{z}_0)$ to a simpler prior distribution $\bm{z}_t$ via a diffusion process:
		\vspace{-3pt}
	\begin{equation}\small
		\bm{z}_t\sim \mathcal{N}\left(\sqrt{\alpha_t}\bm{z}_{t-1}, (1-\alpha_t)\bm{I}\right),
			\vspace{-3pt}
	\end{equation}
	where $\{\alpha_t\}$ are fixed coefficients that determine the noise schedule. \textbf{(ii) Reverse pass} leverages a trained neural network $\epsilon_\theta$ (typicall a UNet) to gradually reverse the effects of the forward process by training it to estimate the noise $\epsilon$ which has been added to $\bm{z}_0$ . Hence, the training objective can be derived as: 
	\vspace{-3pt}
	\begin{equation}\small
		\mathcal{L}_\text{DM} \!=\! \mathbb{E}_{\bm{z}_0, \epsilon\sim\mathcal{N}(0,1), t} \left[ ||\epsilon - \epsilon_\theta (\bm{z}_t(\bm{z}_0, \epsilon), t; \mathcal{C})||_2^2 \right],
			\vspace{-3pt}
	\end{equation}
	where $\mathcal{C}$ denotes an additional conditioning input to $\epsilon_\theta$. Further, latent diffusion models (LDMs) extend DMs by training them in the low-dimensional latent space of an autoencoding model $\mathcal{E}$ (\eg, VQGAN \cite{esser2021taming}):
		\vspace{-3pt}
	\begin{equation}\small
		\mathcal{L}_\text{LDM} \!=\! \mathbb{E}_{\mathcal{E}(\bm{z}_0), \epsilon\sim\mathcal{N}(0,1), t} \left[ ||\epsilon - \epsilon_\theta (\bm{z}_t(\mathcal{E}(\bm{z}_0), \epsilon), t; \mathcal{C})||_2^2 \right].
			\vspace{-3pt}
	\end{equation}
	This leads to many popular text-to-image DMs (T2I-DMs), \ie, Stable Diffusion (SD) \cite{rombach2022high}. Current T2I-DMs are able to generate high-fidelity images with rich texture, diverse content and intricate structures while having compositional and editable semantics. This phenomenon potentially suggests that T2I-DMs can implicitly learn both high-level and low-level visual concepts from massive image-text pairs. Moreover, recent research  has highlighted the clear correlations between attention maps and  text prompts in T2I-DMs \cite{hertz2022prompt,parmar2023zero}. These properties extend the capability of T2I-DMs from generation to  perception tasks  \cite{li2023your,clark2024text}.

	\noindent~$\bullet$~\textbf{\textit{Self-Distillation with No Labels (DINO\&DINOv2).}} DINO \cite{caron2021emerging} interprets self-supervised learning of ViTs as a special case of self-distillation, wherein learning relies on model's own predictions rather than external labels.
	Despite being a relatively \textit{small-sized} model, DINO demonstrates a profound understanding of the visual world, characterized by its highly structured feature space. Notably, DINO shows two emerging properties: its features are excellent k-NN classifiers, and  contain explicit information pertaining to image segmentation.  DINOv2 \cite{oquab2024dinov2}  pushes the limits of visual features by scaling DINO in model and data sizes, along with an improved training recipe. The resultant model yields general-purpose features that close the performance gap with supervised alternatives across various benchmarks, while also showing notable properties, such as understanding of object parts and scene geometry. \textit{Strictly, speaking, DINO is not a `large' model in terms of the parameter scale, but it is included due to the emerged nice properties for segmentation, and its role as the  successor of DINOv2 .}

	\noindent~$\bullet$~\textbf{\textit{Segment Anything (SAM)}}. SAM \cite{kirillov2023segment} has sparked a revolution in the field of image segmentation, and profoundly influences the development of large, general-purposed  models in computer vision. Unlike the aforementioned vision FMs, SAM is built specifically for image segmentation, which is trained on a corpus of 1 billion masks from 11 million images using a promptable segmentation task. It achieves powerful zero-shot task generality to handle a wide range of image segmentation tasks, and allows for enhanced interactivity in segmentation through the use of ``prompts'' in  forms of points, masks, boxes, and even language. Beyond this, SAM has shown promising capabilities in a multitude of tasks, including medical imaging \cite{ma2024segment}, image editing \cite{xie2023edit}, video  segmentation \cite{cheng2023segment}. Despite its capabilities, one downside of SAM is the computational expense associated with its heavy image encoder. However, SAM2 \cite{ravi2024sam} addresses this by  instead using an MAE pre-trained Hiera as the image encoder, yielding real-time speed and improved  segmentation accuracy.

	\section{Segmentation Knowledge Emerges from FMs}\label{sec:3}
\begin{figure}[!htb]
		\vspace{-2pt}
		\centering
		\includegraphics[width=0.99\linewidth]{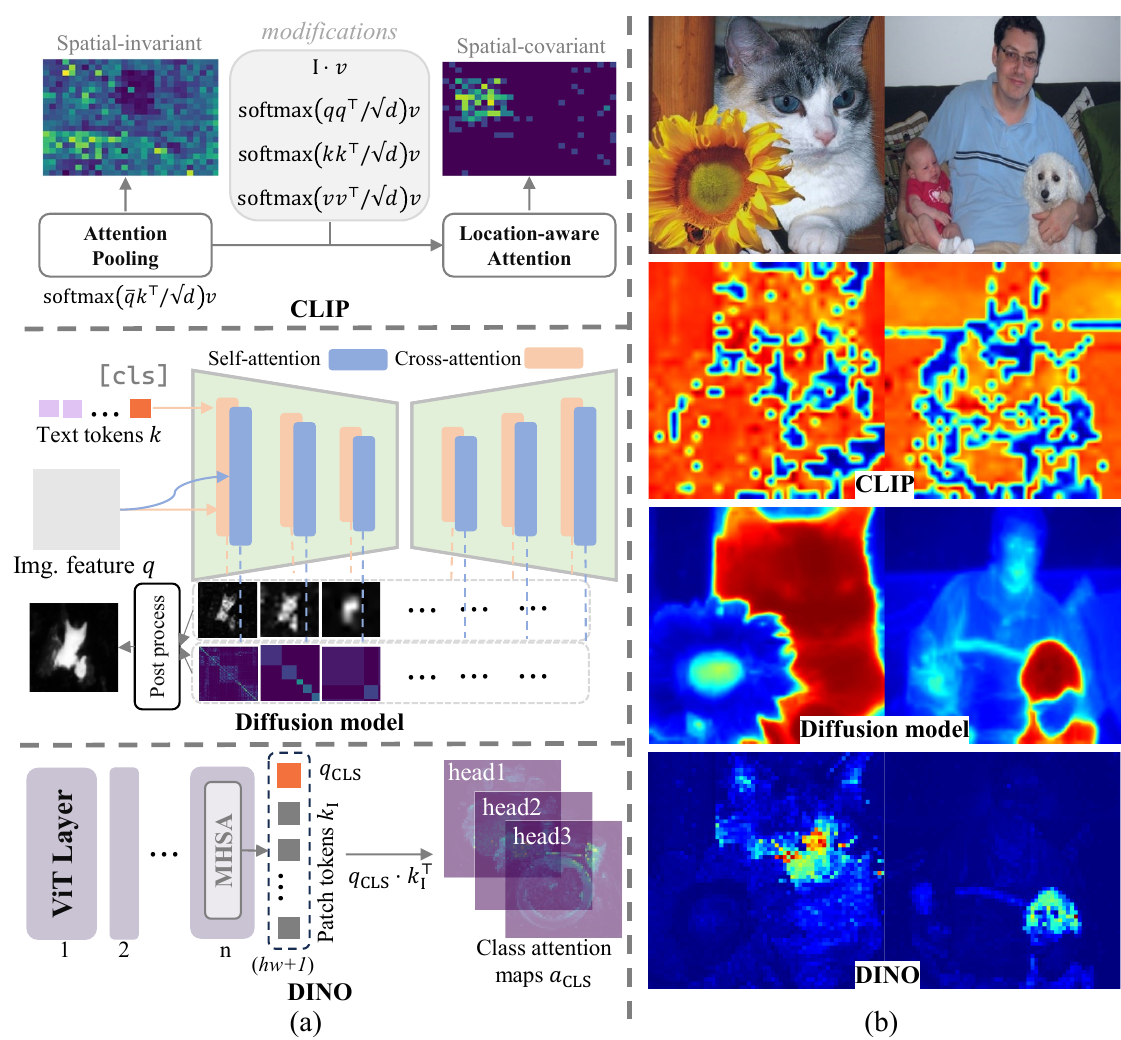}
		\vspace{-10pt}
		\caption{ (a) Illustrations of how segmentation derived from FMs. Briefly speaking, Modifying CLIP's attention pooling to location-aware attentions can obtain segmentation features. Merging cross-attention maps and self-attention maps in DMs can produce precise semantic segments. DINO naturally contains segmentation properties in the last attention maps of the class token.  (b) shows  some visualization examples.
		}
		\vspace{-12pt}
		\label{fig:3}
	\end{figure}

	Given the  emergency capabilities of LLMs, a natural question arises: \textit{Do segmentation properties emerge from  FMs?} The answer is positive, even for FMs not explicitly designed for segmentation, such as CLIP, DINO and Diffusion Models.   In this section, we elaborate on the techniques to extract segmentation knowledge from these FMs, which are effectively unlocking a new frontier in image segmentation, \ie,  acquiring segmentation without any training. {Fig.~\ref{fig:3} illustrates how to approach this and shows some examples.}
	
	\subsection{Segmentation Emerges from CLIP}  \label{sec:3.1}
	
	Many studies \cite{zhou2022extract,wang2023sclip,ranasinghe2023perceptual}  acknowledge that the standard CLIP is able to discern the appearance of objects, but is limited in  understanding their locations. The main reason is that CLIP  learns holistic visual representations that are invariant to spatial positions, whereas segmentation requires spatial-covariant features -- local representations should vary \textit{w.r.t.} their spatial positions in an image. To better explain this, we revisit  self-attention in Transformers:
	\vspace{-3pt}
	\begin{equation}\small\label{eq:selfattention}
		\text{SelfAttention}(\bm{q}, \bm{k}, \bm{v}) = \underbrace{\text{softmax}\left(\bm{q}\bm{k}^\top/\sqrt{d}\right)}_{\text{self-attention matrix}}\bm{v},
			\vspace{-3pt}
	\end{equation}
	where  $\bm{q} \!=\! {\bm{x}}\bm{W}_q \!\in\!\mathbb{R}^{N\times d}$, $\bm{k} \!=\! \bm{x}\bm{W}_k \!\in\!\mathbb{R}^{N\times d}$, $\bm{v} \!=\! \bm{x}\bm{W}_v \!\in\!\mathbb{R}^{N\times d}$ are query, key, and value embeddings. $\bm{x}\!\in\!\mathbb{R}^{N\times d}$ is the input sequence with $N$ tokens, each being a $d$-dimensional vector. $\bm{W}\!\in\!\mathbb{R}^{d\times d}$ denotes a projection matrix  whose parameters are learned in pre-training. CLIP applies  attention pooling to the last self-attention layer: 
	\begin{equation}\small\label{eq:attentionpooling}
		\begin{aligned}
			\!\!\!\!\text{AttentionPooling}(\bm{q}, \bm{k}, \bm{v}) = 	\text{SelfAttention}(\bar{\bm{q}}, \bm{k}, \bm{v})
		\end{aligned}
	\end{equation}
	where $\bar{\bm{q}} \!=\! \bar{\bm{x}}\bm{W}_q \!\in\!\mathbb{R}^d$, and $\bm{\bar{x}}\!\in\!\mathbb{R}^d$ is globally average-pooled feature of $\bm{x}$. Eq.~\ref{eq:attentionpooling} encourages similar representations for different locations, leading to spatial-invariant features. 
	
	Despite this, MaskCLIP \cite{zhou2022extract} finds that it is feasible to extract segmentation knowledge from CLIP with minimal modifications to the attention pooling module.  Specifically, it simply sets the attention matrix to an identity matrix. In this way, each local visual token  receives information only from its corresponding position so that visual features (\ie, $\bm{v}$) are well localized. Such a straightforward modification results in an 11\% increase of CLIP's mIoU on COCO-Stuff. Furthermore, SCLIP \cite{wang2023sclip} proposes to compute pairwise token correlations to  allow each local token to attend to  positions sharing similar information, \ie, the attention matrix is computed as: $\left(\text{softmax}(\bm{q}\bm{q}^\top)+\text{softmax}(\bm{k}\bm{k}^\top)\right)$. CLIPSurgery \cite{li2023clip}  computes value-value attention matrix: $\left(\text{softmax}(\bm{v}\bm{v}^\top)\right)$ and incorporates the attention into each Transformer block rather than the last one. NACLIP \cite{hajimiri2024pay} computes key-key attention matrix: $\left(\text{softmax}(\bm{k}\bm{k}^\top)\right)$ and further weights the attention map with a  Gaussian kernel to encourage  more consistent attention across adjacent patches. GEM \cite{bousselham2024grounding} presents a generalized way to calculate the attention matrix as: $\left(\text{softmax}(\bm{q}\bm{q}^\top)+\text{softmax}(\bm{k}\bm{k}^\top)+\text{softmax}(\bm{v}\bm{v}^\top)\right)$.
	
	\subsection{Segmentation Emerges from DMs}  \label{sec:3.2}
	A family of methods \cite{wang2023diffusion,yoshihashi2023attention,nguyen2024dataset,tang2023daam,corradini2024freeseg,marcos2024open,kawano2024maskdiffusion,karazija2023diffusion,zhou2024prototype,wang2023visual} shows that pre-trained generative models, especially DMs, manifest remarkable segmentation capabilities.  A major insight is that segmentation emerges from cross-attention maps in DMs. Formally, the cross-attention at one layer is computed as:
	\begin{equation}\small
		\begin{aligned}
			& \bm{m}  = \text{CrossAttention}(\bm{q}, \bm{k}) = \text{softmax}(\bm{q}\bm{k}^{\top}/\sqrt{d}), \\
			&\text{where}~~~\bm{q} = \Phi(\bm{z}_t) \in \mathbb{R}^{hw\times d}, ~\bm{k} = \Psi(\bm{e}) \in \mathbb{R}^{N\times d}.
		\end{aligned}
	\end{equation}
	Here $\Phi$ and $\Psi$ indicate linear layers of the U-Net  that denoise in the latent space. $N$ and $d$ represent the length of text tokens and feature dimensionality in the layer, respectively. $hw$ is the spatial size of the feature. $\bm{m}\!\in\!\mathbb{R}^{hw\times N}$ denotes the cross-attention map of a single head. As seen, $\bm{m}$ captures dense correlations between pixels and words, from which we are able to extract the mask $\bm{m}_\texttt{CLS}\!\in\!\mathbb{R}^{hw}$ associated with the class token $[\texttt{CLS}]$. In practice, most methods consolidate cross-attention matrices across blocks, timestamps, and attention heads into a single attention map \cite{wang2023diffusion,yoshihashi2023attention,nguyen2024dataset,tang2023daam,kawano2024maskdiffusion} to obtain higher-quality attention maps. Nevertheless, cross-attention maps often  lack clear object
	boundaries and may exhibit internal holes. Thus, they are typically completed by incorporating self-attention maps \cite{wang2023diffusion,nguyen2024dataset} to yield final segmentation mask as $\hat{\bm{m}}_\texttt{CLS}=\bm{a}_\texttt{SA}\bm{m}_\texttt{CLS}$ where $\bm{a}_\texttt{SA}\!\in\!\mathbb{R}^{hw\!\times\!hw}$ is a self-attention matrix. 

	\subsection{Segmentation Emerges from DINO}  \label{sec:3.3}
	DINO \cite{caron2021emerging} and DINOv2 \cite{oquab2024dinov2} demonstrate a surprising phenomenon that segmentation knowledge emerges in \textit{self-supervised} visual transformers, but not appear explicitly in either supervised ViTs or CNNs.   Caron \textit{et al.} show in DINO \cite{caron2021emerging} that sensible object segmentation can be obtained from the self-attention of  class token \texttt{[CLS]}  in the last attention layer. More formally, given an  input sequence of $M$ ($=hw$) patches, the affinity vector $\bm{\alpha}_\texttt{CLS}$ can be computed as the pairwise similarities between the class token \texttt{[CLS]}  and patch tokens \texttt{[I]} in an attention head of the last layer:
	\begin{equation}\small\label{eq:cls2patch}
		\bm{\alpha}_\texttt{CLS} = \bm{q}_\texttt{CLS} \cdot \bm{k}_\texttt{I}^\top ~~~\in\mathbb{R}^{1\times M},
	\end{equation}
	where $\bm{q}$ and $\bm{k}$ denote \textit{query} and \textit{key} features of corresponding tokens, respectively. The final attention map are averaged of $\bm{\alpha}_\texttt{CLS}$ over all attention heads, and can directly binarized to yield  segmentation masks. 
	
	Beyond this, some other works \cite{simeoni2021localizing,melas2022deep,van2022discovering,zadaianchuk2023unsupervised}  localize objects based on  similarities between patch tokens:
	\begin{equation}\small\label{eq:affinity}
		\bm{A}_\texttt{I} = \bm{k}_\texttt{I} \cdot \bm{k}_\texttt{I}^\top ~~~\in\mathbb{R}^{M\times M}.
	\end{equation}
	Here each element in $\bm{A}_\texttt{I}$ measures the similarity between a pair of tokens. The \textit{key} features are typically chosen in the computation since they show better localization properties than others (\ie, \textit{query} or \textit{value} features) \cite{amir2021deep}. Based on the derived affinity matrix $\bm{A}_\texttt{I}$, LOST \cite{simeoni2021localizing} directly mines potential objects based on an inverse selection strategy; DeepSpectral \cite{melas2022deep} and COMUS \cite{zadaianchuk2023unsupervised} group pixels by partitioning the affinity matrix based on spectral theory; MaskDistill \cite{van2022discovering} selects discriminative tokens based on $\bm{\alpha}_\texttt{CLS}$, and diffuses information of discriminative tokens based on $\bm{A}_\texttt{I}$ to estimate initial segmentation results.

	\begin{figure}[t]
	\centering
	\includegraphics[width=0.99\linewidth]{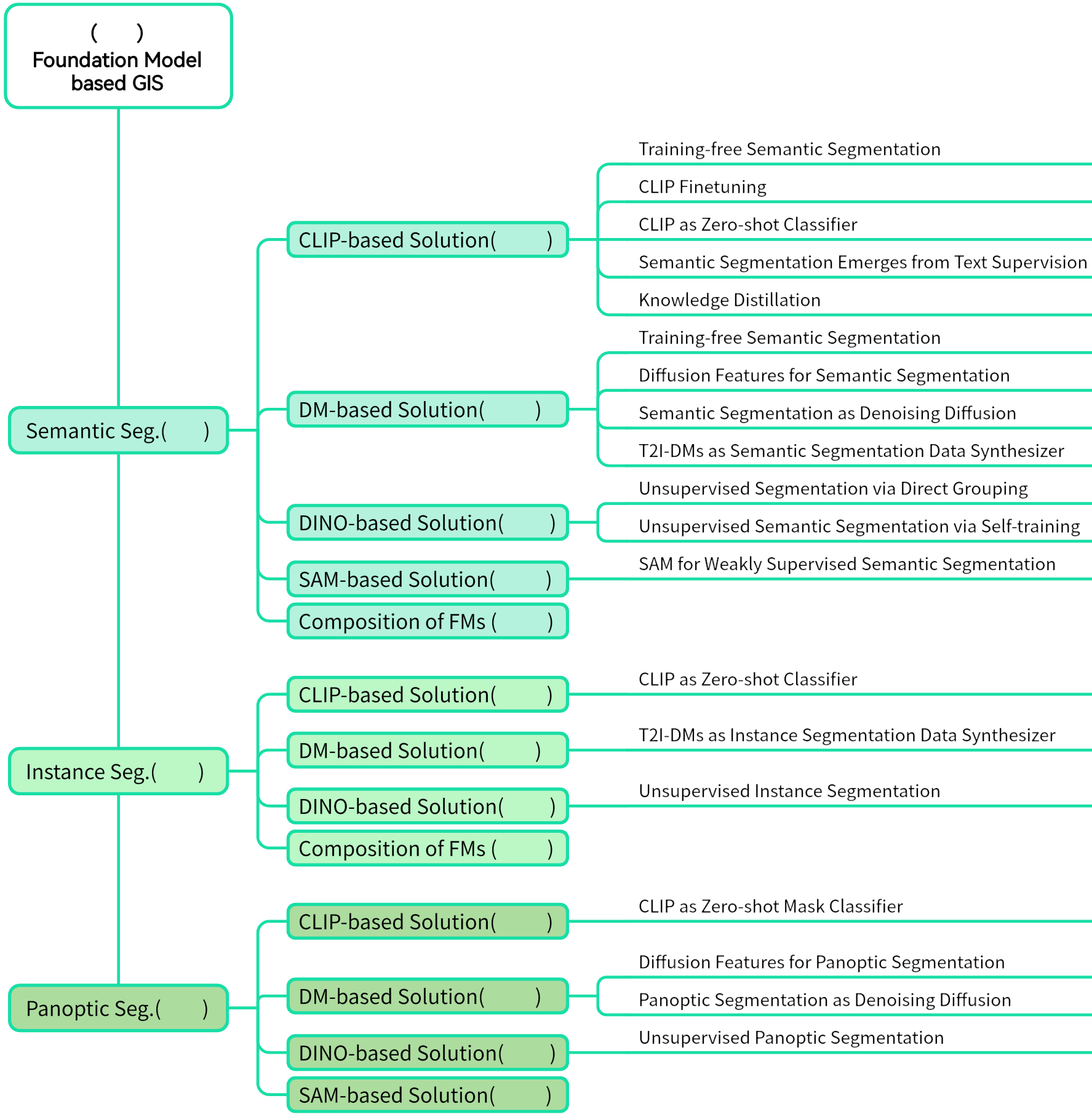}
        \put(-226.5, 246.8){\tiny \S\ref{sec:4}}
        \put(-212.3, 155.7){\tiny \S\ref{sec:4.1}}
        \put(-136, 199){\tiny \ref{sec:4.1.1}}
        \put(-138.5, 160.4){\tiny \ref{sec:4.1.2}}
        \put(-135, 134.5){\tiny \ref{sec:4.1.3}}
        \put(-137, 121.6){\tiny \ref{sec:4.1.4}}
        \put(-136, 112){\tiny \ref{sec:4.1.5}}
        \put(-214.1, 77.5){\tiny \S\ref{sec:4.2}}
        \put(-136, 95){\tiny \ref{sec:4.2.1}}
        \put(-139, 82.8){\tiny \ref{sec:4.2.2}}
        \put(-135, 70){\tiny \ref{sec:4.2.3}}
        \put(-136, 60){\tiny \ref{sec:4.2.4}}
        \put(-213.5, 23.5){\tiny \S\ref{sec:4.3}}
        \put(-136, 43.7){\tiny \ref{sec:4.3.1}}
        \put(-139, 26.3){\tiny \ref{sec:4.3.2}}
        \put(-135, 13.3){\tiny \ref{sec:4.3.3}}
        \put(-135.5, 3.3){\tiny \ref{sec:4.3.4}}
	\caption{Overview of Foundation Model based GIS (\S\ref{sec:4}).}
	\label{fig:gis}
        \end{figure}

	\section{Foundation Model based GIS}\label{sec:4}
	
	This section presents a comprehensive review of recent advances in FM-based GIS, including semantic (\S\ref{sec:4.1}), instance (\S\ref{sec:4.2}) and panoptic segmentation (\S\ref{sec:4.3}), as illustrated in Fig.~\ref{fig:gis}. Our discussions are approached from a \textit{technical} perspective  to elucidate the fundamental concepts and highlight the roles of FMs in GIS.
	
	\subsection{Semantic Segmentation}\label{sec:4.1}
	
	\subsubsection{CLIP-based Solution}\label{sec:4.1.1}
	
	\textit{How to transfer pre-trained knowledge in CLIP to segmentation?} This question has driven a wide spectrum of studies to solve image segmentation based on CLIP. However, the task is challenging due to the inherent granularity gap  between the image-level training task in CLIP and  pixel-level prediction task in image segmentation. Popular solutions are:
	
	\noindent~$\bullet$~\textbf{\textit{Training free Semantic Segmentation.}} As discussed in \S\ref{sec:3.1}, it is feasible to derive segmentation masks from CLIP,  with a slight modification of the self-attention module. On this basis, many approaches \cite{zhou2022extract,wang2023sclip,li2023clip,hajimiri2024pay,bousselham2024grounding} achieve semantic segmentation by leveraging CLIP text encoder as the classifier to determine the category of each mask.  The whole process involves no extra training or fine-tuning.

	\noindent~$\bullet$~\textbf{\textit{CLIP Finetuning.}} Following the popular \textit{pre-training-then-fine-tuning} paradigm, a large number of methods fine-tunes CLIP using segmentation data.  They can be categorized as either \textbf{full fine-tuning} or \textbf{parameter-efficient tuning} approaches. Full fine-tuning methods entail tuning the entire visual or textual encoders of CLIP. DenseCLIP \cite{rao2022denseclip}, for instance, pioneers this approach by refining CLIP's visual encoder through solving a pixel-text matching task.  PPL \cite{kwon2023probabilistic} augments  DenseCLIP with a probabilistic framework to learn more accurate textual descriptions based on visual cues.  Though showing promising results, these methods tend to break the visual-language association within CLIP and lead to severe losses of the open-vocabulary capacity. To alleviate this, CATSeg \cite{cho2024cat} introduces a cost aggregation-based framework to maintain the zero-shot capability of CLIP even after full fine-tuning. OTSeg \cite{kim2024otseg} tackles it by leveraging the ensemble of multiple text prompts, and introduce a multi-prompts sinkhorn attention to improve multimodal alignment. However, these methods typically necessitate a substantial volume of densely annotated training images. In contrast, ZegCLIP \cite{zhou2023zegclip}, LDVC \cite{zhang2024language}, and ZegOT \cite{kim2023zegot} employ  parameter-efficient prompt tuning techniques to transfer CLIP.  To prevent overfitting to seen categories, they all learn image-specific textual embeddings to achieve more accurate pixel-text alignment. SemiVL \cite{hoyer2023semivl}  adopts partial tuning strategies to only tune parameters of self-attention layers. SAN \cite{xu2023side} adapts CLIP image encoder to segmentation via a lightweight adapter, and decouples the mask proposal and classification stage by predicting attention biases applied to deeper layers of CLIP for recognition.

	\noindent~$\bullet$~\textbf{\textit{CLIP as  Zero-Shot Classifier.}} Apart from model fine-tuning, many efforts directly utilize the pre-trained CLIP as  classifiers, and are able to preserve CLIP's zero-shot transferability. The methods can be categorized into two major types: \textbf{mask classification} and \textbf{pixel classification}.  
	
	Mask classification methods \cite{xu2022simple,sun2024clip,ding2022decoupling,han2023global,zhang2023associating,han2023open,gui2024knn,liang2023open,jiao2023learning} in general follow a two-stage paradigm, wherein class-agnostic mask proposals are firstly extracted and then the pre-trained CLIP is used for classifying the proposals. Pioneering studies \cite{xu2022simple,sun2024clip} require a standalone, CLIP-unaware model for proposal generation, while recent approaches \cite{ding2022decoupling,han2023global,han2023open,zhang2023associating,gui2024knn} tend to integrate mask generation and classification within a unified framework. All these methods maintain CLIP frozen during training, but the vanilla CLIP  is insensitive to different mask proposals, constraining classification performance. OVSeg \cite{liang2023open} and  MAFT \cite{jiao2023learning} tackle this issue by tuning CLIP during training to make it more mask-aware.

	Pixel classification methods \cite{li2022language,ma2022fusioner,mukhoti2023open,ranasinghe2023perceptual,lin2023clip,he2023clip} employ CLIP to recognize  pixels.  LSeg \cite{li2022language} achieves this by learning an independent image encoder to align with  the original textual encoder in CLIP. Fusioner \cite{ma2022fusioner} introduces a cross-modality fusion module to capture the interactions between  visual and textual features from the frozen CLIP, and decodes the fused features into segmentation masks. PACL \cite{mukhoti2023open} defines a new compatibility function for contrastive loss to align patch tokens of the vision encoder and the \texttt{[CLS]} token of the text encoder. Patch-level alignment can benefit zero-shot transfer to semantic segmentation.  CLIPpy \cite{ranasinghe2023perceptual} endows CLIP with perceptual grouping with a series of modifications on the aggregation method and pre-training strategies. Due to the absence of fine-grained supervisions,  such CLIP-based segmentors cannot delineate the fine shape of targets. SAZS \cite{liu2023delving} alleviates this by developing a boundary-aware constraint.
	
	\noindent~$\bullet$~\textbf{\textit{Semantic Segmentation Emerges from Text Supervision.}} Inspired by CLIP, a stream of research attempts to learn transferable semantic segmentation models purely from text supervision. GroupViT \cite{xu2022groupvit} and SegCLIP \cite{luo2023segclip} augment  vanilla ViTs with grouping modules to progressively group image pixels into segments. To address their granularity inconsistency issue, SGP \cite{zhang2024uncovering} further mines non-learnable prototypical knowledge \cite{zhou2024prototype} as explicit supervision for group tokens to improve clustering results. Unlike these works require customized image encoders, \cite{yi2023simple} avoids modifying CLIP's architecture, but improves the optimization by sparsely contrasting on the image-text features with the maximum responses. TagAlign \cite{liu2023tagalign} also focuses on the optimization part, and introduces fine-grained attributes as supervision signals to enable dense image-text alignment.
	
	
	\noindent~$\bullet$~\textbf{\textit{Knowledge Distillation (KD).}} KD \cite{hinton2015distilling} is a simple but efficient approach to transfer the capability of a foundation model, which has achieved many  successes in NLP and CV. In the field of semantic segmentation, ZeroSeg \cite{chen2023exploring} and CLIP-ZSS \cite{chen2023clip} distill the semantic knowledge from CLIP's visual encoder to a segmentation model. Moreover, many methods are based on self-distillation to   teach themselves by aligning localized dense feature to visual feature of corresponding image patch \cite{wu2024clipself}, or learning global semantics based on local information \cite{naeem2023silc}. Moreover, CLIP-DINOiser \cite{wysoczanska2023clip} treats DINO as a teacher to guide CLIP learn DINO-like features that are friendly to segmentation.


	\subsubsection{DM-based Solution}\label{sec:4.1.2}
	Beyond the discriminative model  CLIP, there has been a growing interest in extending the horizon of generative models like DMs from generation tasks to semantic segmentation.  
	From the technical perspective, current research can be grouped into the following categories.
	
	\noindent~$\bullet$~\textbf{\textit{Training free Semantic Segmentation.}} Based on the techniques in \S\ref{sec:3.2},  \cite{wang2023diffusion,nguyen2024dataset,tang2023daam} generate a mask $\bm{m}_\texttt{CLS}$ for each candidate class, and assign a category to each pixel by identifying the class with the highest confidence value. FreeSeg-Diff \cite{corradini2024freeseg} follows a two-stage paradigm, that is, cluster attention maps into class-agnostic masks and then classify each mask by CLIP. These methods are limited by text prompt tokens, requiring an association between each semantic class and a prompt word, which is not always valid. To address this, OVAM \cite{marcos2024open} introduces an extra attribution prompt to enable the generation of semantic segmentation masks described by an open vocabulary, irrespective of the words in the text prompts used for image generation. Furthermore, OVDiff \cite{karazija2023diffusion} takes a prototype learning perspective \cite{zhou2024prototype,wang2023visual} to build a set of categorical prototypes using T2I-DMs, which serve as nearest neighbor classifiers for segmentation. DiffSeg \cite{tian2024diffuse} introduces an iterative merging process to merge self-attention maps in SD into valid segmentation masks. Unlike aforementioned methods, FreeDA \cite{barsellotti2024training}  employs SD to build a large pool of visual prototypes, and the most similar prototype is retrieved for each pixel to yield segmentation prediction.

	\noindent~$\bullet$~\textbf{\textit{Diffusion Features for Semantic Segmentation.}} Beyond attention maps, the  harness of DMs' latent representations for semantic segmentation  is  gaining popularity.  Works like \cite{baranchuk2022label,mukhopadhyay2023text} extract internal embeddings from text-free DMs for segmentation, but they are limited to close-vocabulary settings.  In contrast, a majority of methods \cite{zhao2023unleashing,kondapaneni2023text,pnvr2023ld} employs T2I-DMs (mostly SD) to mine semantic representations. LD-ZNet \cite{pnvr2023ld} shows that 1) the latent space of LDMs is a better input representation compared to other forms like RGB images for semantic segmentation, and 2) the middle layers (\ie, \{6,7,8,9,10\}) of the denoising UNet contain more semantic information compared to either the early or later blocks of the encoder (consistent with the observation in \cite{hedlin2024unsupervised}). Beyond this, for T2I-DMs,  text prompt plays a crucial role in feature extraction as it serves as guidance for semantic synthesis.  VPD \cite{zhao2023unleashing} adopts a straightforward method to use class names in the dataset to form the text context of SD, in which class embedding is extracted from the text encoder of CLIP (with prompt ``\texttt{a photo of [CLS]}'').  TADP \cite{kondapaneni2023text} and Vermouth \cite{dong2024bridging} find that automatically generated captions serve as image-aligned text prompt that helps extract more semantically meaningful visual features. In contrast, MetaPrompt \cite{wan2023harnessing} integrates SD with a set of learnable emebddings (called meta prompts) to activate task-relevant features within a recurrent feature refinement process. Furthermore, latent features show exceptional generalization performance to unseen domains with proper prompts.

	\noindent~$\bullet$~\textbf{\textit{Semantic Segmentation as Denoising Diffusion.}} Away from these mainstream battlefields, some works \cite{wolleb2022diffusion,zbinden2023stochastic,rahman2023ambiguous,ji2023ddp} reformulate semantic segmentation as a denoising diffusion process. They learn an iterative denoising process to predict  the ground truth map $\bm{z}_0$ from random noise $\bm{z}_t\sim\mathcal{N}(0, 1)$ conditioned on corresponding visual features derived from an image encoder. Based on this insight, SegRefiner \cite{wang2023segrefiner} considers a discrete diffusion formulation to refine coarse masks derived from existing segmentation models. Moreover, Peekaboo \cite{burgert2022peekaboo} is an interesting approach that treats segmentation as a foreground alpha mask optimization problem which is optimized  via SD at inference time. It alpha-blends an input image with random background to generate a composite image, and then takes an inference time optimization method to iteratively update the alpha mask to converge to  optimal segmentation with respect to image and text prompts. 
	
		\noindent~$\bullet$~\textbf{\textit{T2I-DMs as Semantic Segmentation Data Synthesizer.}} Collecting and annotating images with pixel-wise labels is time-consuming and laborious, and thus always a challenge to semantic segmentation. With recent advances in AIGC, many studies  \cite{nguyen2024dataset,wu2023diffumask,wu2024datasetdm,xie2023mosaicfusion} explore the potential of T2I-DMs to build large-scale  segmentation dataset (including synthetic images and associated mask annotations), which  serve as a more cost-effective data source to train any existing semantic segmentation models. The idea has also been adopted in specialist domains like medical image segmentation \cite{fernandez2022can}. Rather than directly generating synthetic masks, some works \cite{schnell2023generative,yu2023diffusion,zhang2023emit} employ T2I-DMs for data augmentation based on a few labeled images.

	\subsubsection{DINO-based Solution} \label{sec:4.1.3}

	
	\noindent~$\bullet$~\textbf{\textit{Unsupervised Segmentation via Direct Grouping.}} Given the emergence of segmentation properties in DINO, many methods directly group DINO features into distinct regions via, \eg, k-means \cite{amir2021deep} or graph partition \cite{simeoni2021localizing,wang2022self,wang2023tokencut} based on spatially local affinities in Eq.~\ref{eq:affinity}. While being training-free, they are limited in discovering salient objects, and fail to generate masks for multiple semantic regions -- which is critical for semantic segmentation.

	\noindent~$\bullet$~\textbf{\textit{Unsupervised Semantic Segmentation via Self-training.}}
	Follow-up works investigate \emph{self-training} approaches to address aforementioned limitation. They tend to train segmentation models on automatically discovered pseudo labels from DINO features. Pseudo labels are in general obtained in a bottom-up manner, but the strategies differ across methods. DeepSpectral \cite{melas2022deep}  performs spectral clustering over dense DINO features to over-cluster each image into segments, and afterwards cluster DINO representations of such segments across images to determine pseudo segmentation labels. Those segments represent object parts that could be combined with over-clustering and community detection to enhance the quality of pseudo masks \cite{ziegler2022self}. COMUS \cite{zadaianchuk2023unsupervised} combines unsupervised salient masks with DINO feature clustering to yield initial pseudo masks, which are exploited to train a semantic segmentation network to  self-bootstrap the system on images with multiple objects. Notably, STEGO \cite{hamilton2022unsupervised} finds that DINO's features have correlation patterns that are largely consistent with true semantic labels, and accordingly presents a novel contrastive loss to distill unsupervised DINO features into compact semantic clusters. Furthermore, DepthG \cite{sick2024unsupervised} incorporates spatial information in the form of depth maps into the STEGO training process; HP \cite{seong2023leveraging} proposes more effective hidden positive sample to enhance contrastive learning; EAGLE \cite{kim2024eagle} extracts object-level semantic and structural cues from DINO features to guide the model learning object-aware representations. 
	
	\subsubsection{SAM-based Solution} \label{sec:4.1.4}
	\noindent~$\bullet$~\textbf{\textit{SAM for Weakly Supervised Semantic Segmentation.}} While SAM is semantic unawareness, it attains generalized and remarkable segmentation capability, which are widely leveraged to improve segmentation quality in the weakly supervised situations. \cite{chen2023segment} uses SAM for post-processing of segmentation masks, while \cite{chen2023weakly} leverages SAM for zero-shot inference. S2C \cite{kweon2024sam} incorporates SAM at both feature and logit levels. It performs prototype contrastive learning based on SAM's segments, and extracts salient points from CAMs for prompting SAM.

	\subsubsection{Composition of FMs for Semantic Segmentation} \label{sec:4.1.5}
	
	FMs are endowed with distinct capabilities stemming from their pre-training objectives. For example, CLIP excels in semantic understanding, while SAM and DINO specialize in spatial understanding. As such, many approaches amalgamate an assembly of these FMs into a cohesive system that absorbs their expertise. Some of them are built under zero guidance \cite{rewatbowornwong2023zero,corradini2024freeseg,kawano2024tag}. They leverage DINO or SD to identify class-agnostic segments, map them to CLIP's latent space, and translate the embedding of each segment into a word (\ie, class name) via image captioning models like BLIP. Another example is SAM-CLIP \cite{wang2023sam} that combines SAM and CLIP into a single model via multi-task distillation. Recently, RIM \cite{wang2024image} builds a training-free framework under the collaboration of three  VFMs. Concretely, it first constructs category-specific reference features based on SD and SAM, and then matches them with region features derived from SAM and DINO via relation-aware ranking.

	%
	%
	%
	
	\subsection{Instance Segmentation}\label{sec:4.2}

	\subsubsection{CLIP-based Solution}\label{sec:4.2.1}

	\noindent~$\bullet$~\textbf{\textit{CLIP as Zero-shot  Instance Classifier.}}
	CLIP plays an important role in achieving open-vocabulary instance segmentation. \cite{ding2023open,xu2023masqclip,yu2024convolutions} leverage the frozen CLIP text encoder as a classifier of instance mask proposals.  OPSNet \cite{chen2023open} utilizes CLIP visual and textual embeddings to enrich instance features, which are subsequently classified by the CLIP text encoder. \cite{he2023primitive} introduces a generative model to synthesize unseen features from CLIP text embeddings, thereby bridging semantic-visual spaces and address the challenge of lack of unseen training data. \cite{he2023semantic} presents a dynamic classifier to project CLIP text embedding to image-specific visual prototypes, effectively mitigating  bias towards seen categories as well as multi-modal domain gap.
	
	\subsubsection{DM-based Solution}\label{sec:4.2.2}
	
	\noindent~$\bullet$~\textbf{\textit{T2I-DMs as Instance Segmentation Data Synthesizer.}} DMs play a crucial role in instance segmentation by facilitating the generation of large-scale training datasets with accurate labels. MosaicFusion \cite{xie2023mosaicfusion} introduces a training-free  pipeline that simultaneously generates
	synthetic images via T2I-DMs and corresponding masks through aggregation over cross-attention maps. \cite{ge2022dall} adopts a cut-and-paste approach for data augmentation, where both foreground objects and background images are generated using DMs.  DatasetDM \cite{wu2024datasetdm} presents a semi-supervised approach that first learns a perception decoder to annotate images based on a small set of labeled data, and then generates images and annotations for various dense prediction tasks.

	\subsubsection{DINO-based Solution}\label{sec:4.2.3}
	
	\noindent~$\bullet$~\textbf{\textit{Unsupervised Instance Segmentation.}} Some methods \cite{wang2023cut,cao2024hassod,van2022discovering,arica2024cuvler} attempt to amplify the innate localization abilitiy of DINO to train instance-level segmentation models without any human labels. They typically work in a two-stage \textit{discover-and-learn} process: discover multiple object masks from DINO features by, \eg, recursively applying normalized cuts \cite{wang2023cut}, and then leverage them as pseudo labels to train instance segmentation models. 
	
	\subsubsection{Composition of FMs for Instance Segmentation}\label{sec:4.2.4}
	
	X-Paste \cite{zhao2023x} revisits the traditional data boosting strategy, \ie, Copy-Paste, at scale to acquire large-scale object instances with high-quality masks for unlimited categories. It makes full use of FMs to prepare images, \ie, using SD to generate images  and using CLIP to filter Web-retrieved images. Instances in the images are extracted with off-the-shelf segmentors, which are composed with background images to create training samples. DiverGen \cite{fan2024divergen} improves X-Paste by focusing more on enhancing category diversity. It leverages SAM to more accurately extract instance masks. Orthogonal to these studies,  Zip \cite{shi2024the} combines CLIP and SAM to achieve training-free instance segmentation. It observes that clustering on features of CLIP's middle layer is keenly attuned to object boundaries. Accordingly, it first clusters CLIP features to extract segments, then filters them according to boundary and semantic cues, and finally prompts SAM to produce instance masks. 
	
	Moreover, it is easy to directly turn SAM into an instance segmentation model by feeding bounding boxes of instances as prompts \cite{ke2024segment,ren2024grounded}, which can be obtained from  object detectors, \eg, Faster R-CNN \cite{ren2016faster}, Grounding DINO \cite{liu2023grounding}.

	\subsection{Panoptic Segmentation}\label{sec:4.3}
	
	\subsubsection{CLIP-based Solution}\label{sec:4.3.1}
	
	\noindent~$\bullet$~\textbf{\textit{CLIP as  Zero-Shot Mask Classifier.}} 
	Most recent panoptic segmentation approaches \cite{ding2023open,xu2023masqclip,zou2023generalized,chen2023open,he2023primitive,wang2024open,qin2023freeseg,yu2024convolutions} follow the query-based mask classification framework introduced  in MaskFormer \cite{cheng2021per} / Mask2Former \cite{cheng2022masked}. They generate class-agnostic mask proposals first and then utilize CLIP to classify the proposals, thereby empowering MaskFormer and Mask2Former open-vocabulary segmentation capabilities. MaskCLIP \cite{ding2023open} introduces a set of mask class tokens to extract mask representations more efficiently.  MasQCLIP \cite{xu2023masqclip} augments MaskCLIP by applying additional projections to mask class tokens to obtain optimal attention weights. OPSNet \cite{chen2023open} learns more generalizable mask representations based on CLIP visual encoder that are subsequently used to enhance query embeddings. Unpair-Seg \cite{wang2024open} presents a weakly supervised framework that allows the model to benefit from cheaper image-text pairs. It learns a feature adapter to align mask representations with text embeddings, which are extracted from CLIP's visual and language encoders respectively. Despite the advances, these methods still require training a separate model for each task to achieve the best performance. Freeseg \cite{qin2023freeseg} and DaTaSeg \cite{gu2023dataseg} design all-in-one models with the same architecture and inference parameters to establish remarkable performance in open-vocabulary semantic, instance, and panoptic segmentation problems. OMG-Seg \cite{li2024omg} introduces a unified query representation to unify different task outputs, and is able to handle 10  segmentation tasks across different datasets.
	
	\subsubsection{DM-based Solution}\label{sec:4.3.2}
	
	\noindent~$\bullet$~\textbf{\textit{Diffusion Features for Panoptic Segmentation.}} ODISE \cite{xu2023open} explores internal representations within T2I DMs to accomplish open-vocabulary panoptic segmentation. It  follows the architectural design of Mask2Former but leverages visual features derived from pre-trained diffusion UNet to predict binary mask proposals and associated mask  representations. These proposals are finally recognized using CLIP as the zero-shot classifier.
	
	\noindent~$\bullet$~\textbf{\textit{Panoptic Segmentation as Denoising Diffusion.}} Pix2Seq-$\mathcal{D}$ \cite{chen2023generalist} formulates panoptic segmentation as a discrete data generation problem conditioned on pixels, using a Bit Diffusion generative model \cite{chen2023analog}. DFormer \cite{wang2023dformer} introduces a diffusion-based mask classification scheme that learns to generate mask features and attention masks from noisy mask inputs.  Further, LDMSeg \cite{van2024simple} solves generative segmentation based on SD by first compressing segmentation labels to compact latent codes and then denoising the latents following the diffusion schedule.

	\subsubsection{DINO-based Solution}\label{sec:4.3.3}
	
	\noindent~$\bullet$~\textbf{\textit{Unsupervised Panoptic Segmentation.}} Based on the successes of STEGO \cite{hamilton2022unsupervised} in semantic segmentation and CutLER \cite{wang2023cut} in instance segmentation, U2Seg \cite{niu2024unsupervised} automatically identify ``things'' and ``stuff'' within images to create pseudo labels  that are subsequently used to train a panoptic segmentation model, such as Panoptic Cascade Mask R-CNN  \cite{kirillov2019panoptic}. Moreover, \cite{vodisch2024good} follows the  bottom-up architecture of \cite{cheng2020panoptic} to separately predict semantic  and boundary maps, which are later fused to yield a panoptic segmentation mask.
	
	\subsubsection{SAM-based Solution}\label{sec:4.3.4}
	\noindent~$\bullet$~\textbf{\textit{Towards Semantic-Aware SAM.}} While SAM shows strong zero-shot  performance, its outputs are semantic-agnostic. This drives many research efforts, \eg, Semantic-SAM \cite{li2024semantic}, SEEM \cite{zou2024segment}, to enhance the semantic-awareness of SAM. In addition to visual prompts in SAM for interactive segmentation, these models learn generic object queries to achieve generic segmentation in both semantic and instance levels. In addition, the models are generally trained on a combination of multiple datasets with semantic annotations, such as COCO \cite{lin2014microsoft}, ADE20K \cite{zhou2017scene}, PASCAL VOC \cite{everingham2015pascal}.

	\begin{figure}[t]
		\centering
		\includegraphics[width=0.99\linewidth]{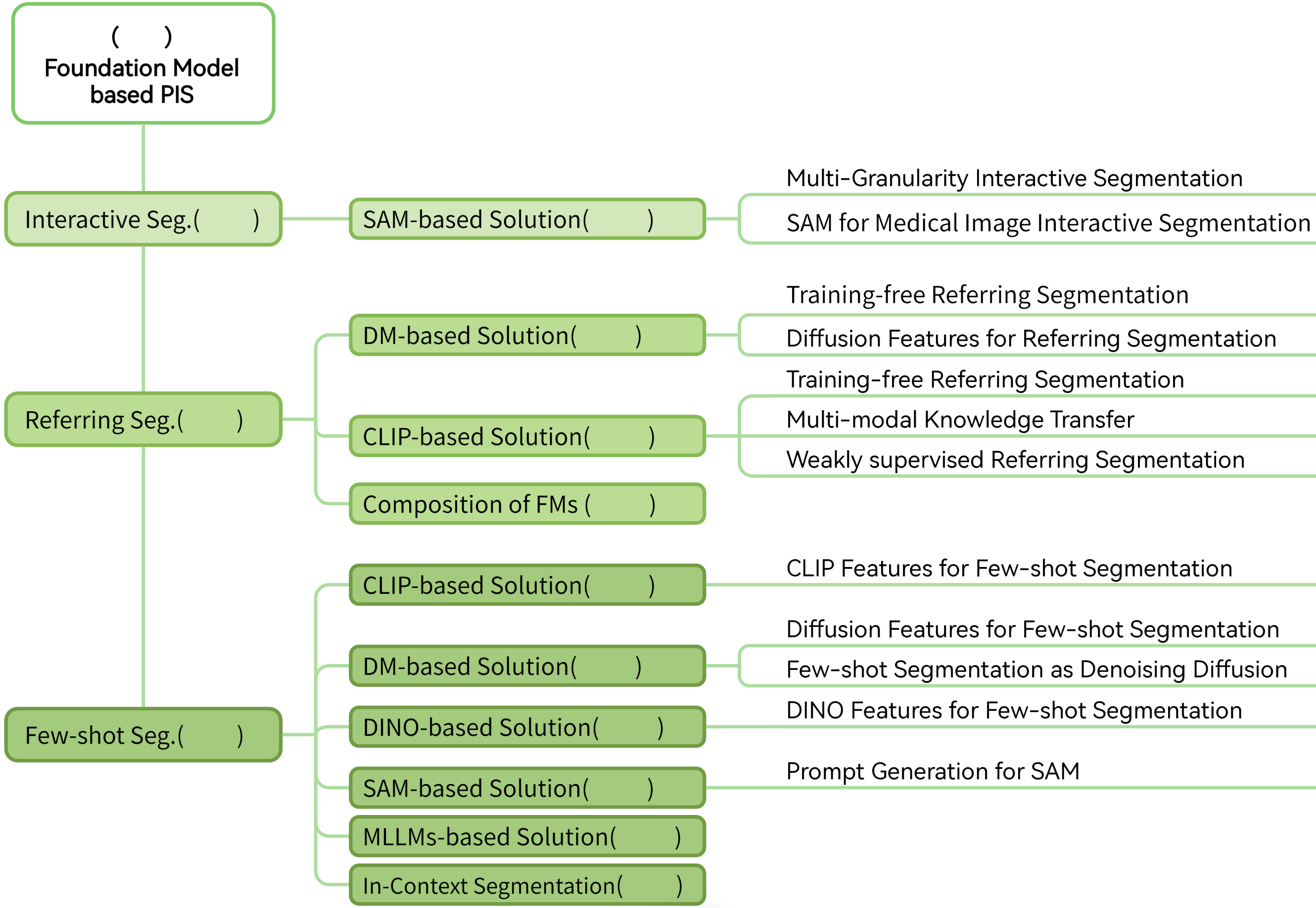}
            \put(-227, 164){\tiny \S\ref{sec:5}}
            \put(-212, 129){\tiny \S\ref{sec:5.1}}
            \put(-137.5, 129){\tiny \ref{sec:5.1.1}}
            \put(-215, 91){\tiny \S\ref{sec:5.2}}
            \put(-140, 107){\tiny \ref{sec:5.2.1}}
            \put(-138, 87.7){\tiny \ref{sec:5.2.2}}
            \put(-138, 75){\tiny \ref{sec:5.2.3}}
            \put(-215, 31){\tiny \S\ref{sec:5.3}}
            \put(-138, 59.5){\tiny \ref{sec:5.3.1}}
            \put(-140, 44){\tiny \ref{sec:5.3.2}}
            \put(-136, 32){\tiny \ref{sec:5.3.3}}
            \put(-138, 20.5){\tiny \ref{sec:5.3.4}}
            \put(-133., 11.5){\tiny \ref{sec:5.3.5}}
            \put(-132, 2.1){\tiny \ref{sec:5.3.6}}
		\caption{Overview of Foundation Model based PIS (\S\ref{sec:5}).}
		\label{fig:pis}
	\end{figure}

	\section{Foundation Model based PIS}\label{sec:5}
	
	As shown in Fig.~\ref{fig:pis}, this section reviews FM-based PIS methods. 
	
	\subsection{Interactive  Segmentation}\label{sec:5.1}
	\subsubsection{SAM-based Solution}\label{sec:5.1.1}
	As SAM  is born as a universe interactive segmenting system, it naturally becomes the top selection for researchers to build advanced interactive segmentation frameworks.
	
	\noindent~$\bullet$~\textbf{\textit{Multi-Granularity Interactive Segmentation.}} Most existing interactive segmentation methods determines a single segmentation mask based on users' input, which ignores spatial ambiguity. In contrast, SAM  introduces a multi-granularity interactive segmentation pipeline, \ie, for each user interaction, desired segmentation region may be the concept of objects with different parts nearby. To  improve the segmentation quality, HQ-SAM~\cite{ke2024segment} proposes a lightweight high-quality output token replace the original SAM’s output token.  After training  on 44K
	highly-accurate masks, HQ-SAM significantly boosts the mask prediction quality of SAM.  Since SAM is class-agnostic, a line of works \cite{pan2024tokenize,yuan2024open} tunes SAM by aligning the query-segmented regions with  corresponding textual representations from CLIP. OVSAM \cite{yuan2024open} explores dual knowledge transfer between SAM and CLIP to enhance SAM's recognition capabilities. Semantic SAM~\cite{li2024semantic} designs a SAM-like framework that supports multi-granularity segmentation  using the captioned SAM data.  Although these multi-granularity interactive segmentation approaches alleviate spatial ambiguity, they result in excessive output
	redundancy and limited scalability. To solve this, GraCo \cite{zhao2024graco} explores  \textit{granularity-controllable} interactive segmentation, which allows precise control of prediction granularity to resolve ambiguity.
	
	\noindent~$\bullet$~\textbf{\textit{SAM for Medical Image Interactive Segmentation.}} Interaction segmentation is crucial in the medical field \cite{zhou2023volumetric}, such as for achieving highly precise segmentation of lesion regions, or reducing manual efforts in annotating medical data. Unlike the segmentation of natural images, medical image segmentation poses greater challenges due to many intrinsic issues such as structural complexity, low contrast, or inter-order variability. Recently, several studies \cite{mazurowski2023segment,cheng2023sam,zhang2024segment} explore the zero-shot interactive segmentation capabilities in medical imaging. They cover a diverse range of anatomical and pathological targets across different medical imaging modalities, including CT \cite{wald2023sam}, MRI \cite{putz2023segment}, pathological images \cite{deng2023segment}, endoscopic images \cite{wang2023sam}. While these studies indicate that SAM performs comparably to state-of-the-art methods in identifying well-defined objects in certain modalities, it struggles or fails completely in more challenging situations, such as when targets have weak boundaries, low contrast, small size, and irregular shapes. This suggests that directly applying SAM without fine-tuning or re-training to previously unseen and challenging medical image segmentation may result in suboptimal  performance.
	
	To enhance SAM's performance on medical images, some approaches propose to fine-tune SAM on medical images. MedSAM \cite{ma2024segment} curates a large scale dataset containing over one million medical image-mask pairs of 11 modalities, which are used for directly fine-tuning SAM. In contrast,  other methods explore parameter-efficient fine-tuning strategies. SAMed \cite{zhang2023customized} applies LoRA modules to the pre-trained SAM image encoder. SAMFE \cite{feng2023cheap} finds that applying LoRA to the mask decoder yields superior performance in cases with few exemplars. SAM-Med2D \cite{cheng2023sam} enhances the image encoder by integrating learnable adapter layers. MedSA \cite{wu2023medical} adapts SAM to volumetric medical images by introducing Space-Depth Transpose where a bifurcated attention mechanism is utilized by capturing spatial correlations in one branch and depth correlations in another. 3DSAM-Adapter \cite{gong20233dsam} introduces a holistic 2D to 3D adaptation method via carefully designed modification of the entire SAM architecture.

	\subsection{Referring  Segmentation}\label{sec:5.2}
	
	\subsubsection{CLIP-based Solution}\label{sec:5.2.1}
	
	Referring  segmentation aims to segment a referent via a natural linguistic expression. The multi-modal knowledge in CLIP is broadly explored to tackle this multi-modal task.
	
	\noindent~$\bullet$~\textbf{\textit{Training-free Referring Segmentation.}} ZS-RS \cite{yu2023zero} represents a training-free referring image segmentation method that leverages cross-modal knowledge in CLIP. It begins by generating instance-level masks using an off-the-shelf mask generator, then extracts  local-global features of masks and texts from CLIP, and finally  identifies the desired mask based on cross-modal feature similarity. TAS \cite{suo2023text} employs a similar pipeline as ZS-RS, but computes more fine-grained region-text matching scores to pick the correct mask.
	
	\noindent~$\bullet$~\textbf{\textit{Multi-modal Knowledge Transfer}.} Many efforts have been devoted to transfer multi-modal knowledge within CLIP from image-level to pixel-level. A common idea \cite{wang2022cris,xu2023bridging,kim2024extending,nguyen2024improving,yan2023eavl,wang2024unveiling,luddecke2022image,zhou2023text,wang2023barleria} is to introduce a task decoder to fuse CLIP's image and textual features, and train it with text-to-pixel contrastive learning \cite{wang2022cris}. In addition to task decoder, ETRIS \cite{xu2023bridging} and RISCLIP \cite{kim2024extending}  integrate a Bridger module to encourage visual-language interactions at each encoder stage. EAVL \cite{yan2023eavl} learns a set of convolution kernels based on input image and language, and do convolutions over the output of task decoder to predict segmentation masks. UniRES \cite{wang2024unveiling} explores multi-granularity referring segmentation to unify object-level and part-level grounding tasks. TP-SIS \cite{zhou2023text} transfers multi-modal knowledge within CLIP for referring surgical instrument segmentation.


	\noindent~$\bullet$~\textbf{\textit{Weakly Supervised Referring Segmentation.}} Moving towards real-world conditions, some work studies weakly supervised referring segmentation to alleviate the cost on pixel labeling. TSEG \cite{strudel2022weakly} computes patch-text similarities with CLIP and guides the classification objective during training with a multi-label patch assignment mechanism. 	TRIS \cite{liu2023referring} proposes a two-stage  pipeline that extracts  coarse pixel-level maps from  image-text attention maps, which are subsequently used to train a mask decoder.

	\subsubsection{DM-based Solution}\label{sec:5.2.2}
	
	\noindent~$\bullet$~\textbf{\textit{Training-free Referring Segmentation}.} Some works~\cite{ni2023ref, burgert2022peekaboo} find that SD is an implicit referring segmentor with the help of generative process. Peekaboo \cite{burgert2022peekaboo} formulates segmentation as a foreground alpha mask optimization problem with SD, where a fine-grained segmentation map should yield a high-fidelity image generation process. In this way, minimizing the discrepancy between the mask-involved noise and the target noise shall give better textual-aligned pixel representations.  Ref-diff \cite{ni2023ref} first generates a set of object proposals from generative models, and  determines the desired mask based on  proposal-text similarities. 
	
	\noindent~$\bullet$~\textbf{\textit{Diffusion Features for Referring Segmentation}.} With the conditioned textual guidance, the modal-intertwined attention maps (\textit{c.f.} \S\ref{sec:3.2}) could intuitively serve as an initial visual dense representation, which could be used to yield the final segmentation mask. VPD~\cite{zhao2023unleashing} introduces a task-specific decoder to process  encoded features fused from cross-attention maps and multi-level feature maps in U-Net. Meanwhile, LD-ZNet~\cite{pnvr2023ld} injects  attention features into a mask decoder for generating better textual-aligned pixel-level masks. Apart from the attention-based utilization, \cite{iioka2023multimodal,qi2024unigs}  directly feed  side outputs from each intermediate layer of the diffusion U-Net as well as the textual embedding, to a  mask decoder to yield final predictions.   
	

	\subsubsection{LLMs/MLLMs-based Solution}\label{sec:5.2.3}
	The success of LLMs/MLLMs has showcased incredible reasoning ability and can answer complex questions, thereby bringing new possibilities to achieve new pixel reasoning and understanding capabilities. In particularly, LISA \cite{lai2024lisa} studies a new  segmentation task, called \emph{reasoning segmentation}. Different from traditional referring segmentation, the segmentors in this setting are developed to segment the object based on implicit query text involving complex reasoning. Notably,  the query text is not limited to a straightforward reference (\eg, \textit{``the front-runner''}), but a a more complicated description involving complex reasoning
	or world knowledge (\eg, \textit{``who will win the race?''}).  LISA  employs  LLaVA~\cite{liu2024visual} to output a text response based on the input image, text query, and a \texttt{[seg]} token. The embedding for the customized \texttt{[seg]} token is decoded into the segmentation mask via SAM decoder. Afterwards, LISA++ \cite{yang2023improved} promotes LISA to differentiate individuals within the same category and enables more natural conversation in multi-turn dialogue. Based on these works, many efforts have been devoted to promote the reasoning capability and segmentation accuracy. LLM-Seg \cite{wang2024llmseg} proposes using SAM to generate a group of mask proposals that selects the best-suited answer as the final segmentation prediction.  Next-Chat \cite{zhang2023next} adds a \texttt{[trigger]} token that depicts the coordinate of the object box as a supplementary input for MLLM to help generate better masks. Similarly, GSVA \cite{xia2024gsva} introduces a rejection token \texttt{[rej]} to relieve the empty-target case where the object referred to in the instructions does not exist in the image, leading to the false-positive prediction. Except for the functional token incorporation, \cite{wei2024lasagna,yang2024empowering} propose using diverse textual descriptions, such as object attribute and part, to enhance the object-text connection for accurate reasoning results. Regarding reasoning costing, PixelLLM \cite{ren2024pixellm} introduces a lightweight decoder to reduce the computational cost in the reasoning process. Osprey \cite{yuan2024osprey} extends MLLMs by incorporating fine-grained mask regions into language instruction, and delivers remarkable pixel-wise visual understanding capabilities.
	

	\subsubsection{Composition of FMs for Referring Segmentation}\label{sec:5.2.4}
	To enhance the textual representation for pixel-level understanding, some methods use LLMs as the text encoder for obtaining improved textual embedding for modal fusion. Particularly, BERT~\cite{devlin2019bert}, due to its simplicity and practicality,  is nearly the top choice among  works~\cite{yang2022lavt,hu2023beyond,liu2023caris,strudel2022weakly,kim2023shatter,chng2024mask,liu2023gres,liu2023polyformer,zhang2022coupalign,yang2023semantics,shah2024lqmformer,wu2023segment}. Most of them  design a  fusion module to bridge the features between the visual encoder and BERT.  In addition, some works~\cite{zhang2023next,pi2024perceptiongpt,zhu2024llmbind} treat LLM as a multi-modal unified handler, and use Vicuna~\cite{zheng2024judging} to map both image and text into a unified feature space, thereafter generating the segmentation output. With the powerful dialogue capabilities of the GPT-series models~\cite{achiam2023gpt}, some works~\cite{sun2024training,ji2023towards,yu2023interactive} employ ChatGPT to  rewrite descriptions with richer semantics, and encourages finer-grained image-text interaction in referring segmentation model training.  
	
	Apart from textual enhancement using LLMs, SAM~\cite{kirillov2023segment} is widely chosen to provide rich segmentation prior for referring segmentation.  \cite{shang2024prompt} presents a prompt-driven framework to bridge CLIP and SAM in an end-to-end manner through prompting mechanisms.	\cite{huang2024deep} focuses on building  referring segmentors based on a simple yet effective bi-encoder design, \ie, adopting SAM and a  LLM to encode  image and text patterns, respectively, and then fuse the multi-modal features for  segmentation predictions. Such a combination of SAM and LLM, without bells and whistles, could be easily extended to the MLLM case. Therefore, \cite{wu2024f,dai2024curriculum} propose to incorporate CLIP with SAM to improve the multi-modal fusion. Specifically, F-LMM \cite{wu2024f} proposes to use CLIP to encode the visual features, which are decoded by SAM to the predicted segmentation map. PPT \cite{dai2024curriculum} first employs  attention maps of CLIP to compute the peak region as the explicit point prompts, which are directly used to segment the query target. 

	\subsection{Few-Shot  Segmentation}\label{sec:5.3}

	\subsubsection{CLIP-based Solution} \label{sec:5.3.1}
	\noindent~$\bullet$~\textbf{\textit{CLIP Features for Few-Shot Segmentation.}} Adopting CLIP to extract effective visual correlation from the support images to help segmentation inference of the query image has formulated a prevailing pipeline to address FSS, which shall be categorized into two streams based on the usage of CLIP-oriented visual feature. The first class~\cite{jeong2023winclip,wang2024rethinking,shuai2023visual,wang2024language,jia2024embedding,you2024weakly} relies on modelling the feature relationship of support-query images to explicitly segment the query image. WinCLIP \cite{jeong2023winclip} aggregates the multi-scale CLIP-based visual features of the reference and query images to obtain an enhanced support-query correlation score map for pixel-level prediction. \cite{wang2024rethinking,shuai2023visual,wang2024language,jia2024embedding} further refine the score maps  with the query- and support-based self-attention maps. \cite{you2024weakly}  introduces the foreground-background correlation from the support images by crafting proper textual prompts. Another line of works~\cite{zhou2024unlocking,luddecke2022image,han2023partseg} focuses on segmenting the query image regulated by the support-image-generated prototypes, where some metric functions, \eg, cosine similarity, shall be involved for the query-prototype distance calculation. RD-FSS \cite{zhou2024unlocking} proposes to leverage the class description from CLIP text encoder as the textual prototypes, which are then correlated with visual features to dense prediction in a cross-attention manner. Additionally, PartSeg \cite{han2023partseg} aggregates both the visual and textual prototypes to help generate the improved query image pixel-level representation. Here the visual prototypes are obtained through correspondingly pooling the CLIP-based visual feature by the reference  segmentation masks. To further enhance the prototypical representation, \cite{luddecke2022image} use CLIP to generate the visual prototypes from the  masked support images, where only interested object is remained.          
	

	\subsubsection{DM-based Solution}\label{sec:5.3.2}
	\noindent~$\bullet$~\textbf{\textit{Diffusion Features for Few-Shot Segmentation.}} 
	The internal representations of DMs are useful for few-shot segmentation. Specifically, \cite{huang2024few} directly leverages the latent diffusion features at specific time step  as the representations of the support image,  which are  decoded along with the original image via a mask decoder.  On the contrary, DifFSS \cite{tan2023diffss} proposes to synthesize more support-style image-mask pairs using DMs. Building on the invariant mask, the generated support images shall include same mask-covered object yet with diverse background, enriching the support patterns for better query segmentation.

	\noindent~$\bullet$~\textbf{\textit{Few-Shot Segmentation as Denoising Diffusion.}} Some studies~\cite{le2024maskdiff,shen2024segicl} tackle few-shot segmentation by solving a denoising diffusion process. They fine-tune SD to explicitly generate  segmentation mask for query images, with the main difference being the condition applied during the fine-tuning.  MaskDiff \cite{le2024maskdiff} uses  query image and support masked images as the condition, while SegICL \cite{shen2024segicl} merely employs the support/query mask as the condition.

		\subsubsection{DINO-based Solution} \label{sec:5.3.3}
	\noindent~$\bullet$~\textbf{\textit{DINO Features for Few-Shot Segmentation.}} There are some efforts \cite{bensaid2024novel,kappeler2023few,kang2023distilling,anand2023one} exploiting latent representations in DINO/DINOv2 to enhance query and support features.	\cite{bensaid2024novel} directly uses DINOv2 to encode query and support images, and shows that DINOv2 outperforms other FMs, like SAM and CLIP. Based on this, SPINO \cite{kappeler2023few}  employs  DINOv2 for few-shot panoptic segmentation. \cite{kang2023distilling,anand2023one} further mine out  query-support correlations through the cross- and self-attention of token embeddings in DINO, leading to more support-aware segmentation.

	\subsubsection{SAM-based Solution} \label{sec:5.3.4}
	\noindent~$\bullet$~\textbf{\textit{Prompt Generation for SAM.}} Given the provided support image sets, a line of works~\cite{feng2024boosting,zhang2024personalize,zhao2024part,sun2024vrp,he2024apseg} focuses on generating proper prompts for SAM to segment the desired target in the query image. Notably, a majority of them  \cite{feng2024boosting,zhang2024personalize,zhao2024part} propose to generate a group of candidate points as prompts based on the support-query image-level correspondence/similarity, where the support mask, highlighting the query object's semantic, is then used to select the object-oriented prompts.  VRP-SAM \cite{sun2024vrp}  learns a set of visual reference prompts based on query-support correspondence, which are fed into a frozen SAM for segmentation. APSeg \cite{he2024apseg} extends VRP-SAM by exploring multiple support embeddings to generate more meaningful prompts for SAM.
	
	
	\subsubsection{LLM/MLLM-based Solution} \label{sec:5.3.5}
	 There are several trials \cite{zhu2024llafs,meng2023few} in adopting LLM/MLLM to address FSS through instruction design. LLaFS \cite{zhu2024llafs}  maps the fused support-query pattern into the language space, and let a LLM to tell the coordinate description of the desired segmentation mask.     
	 \cite{meng2023few} uses GPT-4 as the task planner to divide FSS into a sequence of sub-tasks based on the support set, subsequently calls vision tools such as SAM and GPT4Vision to predict segmentation masks. 
	 
	 	\subsubsection{In-Context Segmentation}\label{sec:5.3.6}
	 The rapid progress of LLMs leads to an emerging ability to learn in-context from just a few examples \cite{brown2020language,wei2022emergent}. Inspired by this,  researchers are exploring a related concept in computer vision called in-context segmentation (ICS). \textit{From the perspective of segmenting a query image based on a few supports,  ICS can be seen as a sub-task of FSS, but it functions directly on pre-trained models without any task-specific finetuning.} Most ICL-emerged LLMs  are generative models trained through masked language modeling or next token prediction strategies, leading to efforts in ICS that mimic these self-supervised methods. Pioneering work like VPInpainting \cite{bar2022visual} approaches visual in-context learning as image inpainting. It defines \textit{visual prompt} as a grid-like single image containing an input-output example(s) and a query, then trains an inpainting model (via MAE \cite{he2022masked}) to predict the missing parts of an image such that it is consistent with given example(s). With this basis, \cite{sun2023exploring,zhang2024makes,zhang2024instruct} propose to retrieve optimal examples from large datasets as the prompt. Additionally, Painter \cite{wang2023images} and SegGPT \cite{wang2023seggpt} are vision generalists built on in-context learning. They unify various vision tasks within the in-context learning framework by standardizing outputs of core vision tasks. Some other studies~\cite{bai2024sequential,sheng2024towards} focus on creating large vision models by formatting images, like language tokenizer, to a group of sequence as visual sentences, and then perform LLM-like training via next token prediction. Notably, developing these visual autoregressive models requires vast amounts of diverse vision data from varied tasks, \eg, image segmentation, depth estimation. PromptDiffusion \cite{wang2023context} explores in-context learning for diffusion models  by fine-tuning SD to generate the query mask conditioned on the support image-mask pair and the query image.  Matcher \cite{liu2024matcher}  utilizes DINOv2 to locate the target in query images by bidirectional matching, and leverages the coarse location information as the prompts of SAM for segmentation. Tyche \cite{rakic2024tyche} introduces a probabilistic approach to ICS by explicitly modeling training and testing uncertainty, and shows significant potential in medical image segmentation.

	\section{Open Issue and Future Direction}\label{sec:6}
	
	Based on the reviewed research, the field of image segmentation has made tremendous progress in the FM era. Nonetheless, given the high diversity and complexity of segmentation tasks, coupled with the rapid evolution of FMs, several critical directions warrant ongoing exploration.

		\vspace{1mm}
	\noindent~$\bullet$~\textbf{Explaining the Emergence of Segmentation Knowledge in FMs.} Despite that different FMs vary significantly in architectures,  data and training objectives, we observe a consistent emergence of segmentation knowledge from them, which drives the development of impressive training-free segmentation models. However, current methods do not fully explain how these FMs learn to understand pixels, especially how pixels interact with other modalities, like texts in CLIP and Text-to-Image Diffusion Models. This calls for novel explainable techniques to enhance our understanding of pixels in FMs. This is crucial to minimize the negative societal impacts in existing FMs, and will broaden more applications of FMs in diverse visual domains and tasks.
	
	\vspace{1mm}
	\noindent~$\bullet$~\textbf{In-Context Segmentation.} Motivated by the success of in-context learning in the language domain, there has been a growing interest in exploring its potential for vision tasks, such as image segmentation. However, the variability in output representations across vision tasks -- such as the differing formats required for semantic, instance, and panoptic segmentation -- renders ICS a particularly challenging problem. While some progress have been made, current results don't show as high performance as bespoke, especially in difficult tasks like panoptic segmentation. Additionally, the ability to perform segmentation at arbitrary levels of granularity through in-context learning remains an unexplored area. Last, the scale of models employed in ICS is considerably smaller compared to the NLP counterparts like GPT-3, which may be a key factor limiting the performance of ICS. To achieve a breakthrough akin to GPT-3 in the vision domain, it is essential to develop large vision models \cite{bai2024sequential}. This task poses significant difficulties and will require extensive collaboration within the vision community to address issues related to  data,  architecture, and training techniques.
	
	 \vspace{1mm}
	\noindent~$\bullet$~\textbf{Mitigating Object Hallucination in MLLMs-based Models.} Although MLLMs have demonstrated significant success in pixel understanding (\textit{c.f.} \S\ref{sec:5.2.3}), they are prone to the issue of \textit{object hallucination} \cite{li2023evaluating} as LLMs. Here object hallucination refers that a model generates unintended descriptions or captions that contain objects which are inconsistent with or even absent from the target image. This issue  greatly undermines  the reliability of these models in real-world applications. Hence, we advocate for future research in MLLMs-based segmentation to rigorously assess object hallucinations for their models, and to incorporate this issue consideration in the development of segmentation models.

	\vspace{1mm}
	\noindent~$\bullet$~\textbf{Powerful and Scalable Data Engine.} Segmentation data are catalysts for progress in image segmentation. Much of the current success in deep learning based image segmentation  owes to datasets such as PASCAL VOC \cite{everingham2015pascal}, COCO \cite{lin2014microsoft}, Cityscapes \cite{cordts2016cityscapes}, and ADE20K \cite{zhou2017scene}. Nonetheless, scaling up image data is a long-standing challenge and is becoming increasingly critical in the FM era,  which calls for a powerful and scalable segmentation data engine. Recently, SAM \cite{kirillov2023segment} tackles this issue with a data engine that labels images via ``model-in-the-loop'', yielding SA-1B with 11M images and 1B masks. Nevertheless, the engine is limited in realistic image labeling and lacks semantic awareness. A promising direction is to incorporate generative models into the system, which would create a more powerful data engine that can scale to arbitrary levels and is more favorable to data-scarcity scenarios like medical imaging \cite{pan20232d} and satellite imagery \cite{toker2024satsynth}. 
		
    \vspace{1mm}
    \noindent~$\bullet$~\textbf{Diffusion Models as the New Data Source.} Text-to-image diffusion models have been proved feasible to build segmentation datasets by generating pairs of synthetic images and corresponding segmentation masks. However, there exists many challenges. First, existing DMs like SD have difficulties in generating complex scenes, \eg, a crowded street with hundreds of objects, closely intertwined objects. To alleviate this, layout or box conditions, instead of solely text, should be provided to guide the generation. Second, the bias in LAION-5B on which SD was trained, might be transferred to the dataset. This issue can be alleviated by absorbing the advancements in addressing the bias problem in generative models. Third, the domain gap between synthetic and real datasets should be continuously studied. Fourth, current approaches are limited in generating data for the task of semantic segmentation  and a limited number of semantic categories, how to generalize them to generate instance-level segmentation masks and scale up the semantic vocabulary are unsolved.

   \vspace{1mm}
   \noindent~$\bullet$~\textbf{Efficient Image Segmentation Model.} While FM-driven segmentation models exhibit remarkable performance, the majority of methods introduce significant computational overheads, such as heavy image encoders for feature computation and costly fine-tuning processes. These challenges impede the broader applicability and affordability of the models in practical scenarios. Key techniques to be explored include knowledge distillation, model compression, and parameter-efficient tuning. Most existing studies focus on improving the deployment efficiency solely for SAM; yet, attention to other FMs is equally vital.

	\section{Conclusion}\label{sec:7}
	
	In this survey, we provide the first comprehensive review to recent progress of image segmentation  in the foundation model era. We introduce key concepts and examine the inherent segmentation knowledge in existing FMs such as CLIP, Diffusion Models, SAM and DINO/DINOv2. Moreover, we summarize more than 300  image segmentation models for tackling generic  and promptable image segmentation tasks. Finally, we highlight existing research gaps that need to be filled and illuminate  promising avenues for future research. We hope that this survey will act as a catalyst, sparking future curiosity and fostering a sustained passion for exploring the potential of FMs in  image segmentation.

	{
		\bibliographystyle{IEEEtran}
		\bibliography{egbib}
	}
	
	\vfill

\end{document}